\pdfoutput=1

\documentclass[11pt]{article}

\usepackage{emnlp2021}

\usepackage{xr}
\makeatletter
\newcommand*{\addFileDependency}[1]{
  \typeout{(#1)}
  \@addtofilelist{#1}
  \IfFileExists{#1}{}{\typeout{No file #1.}}
}
\makeatother

\newcommand*{\myexternaldocument}[1]{%
    \externaldocument{#1}%
    \addFileDependency{#1.tex}%
    \addFileDependency{#1.aux}%
}

\myexternaldocument{appendix}

\RequirePackage{etoolbox}
\RequirePackage{xcolor}

\definecolor{cb-black}      {RGB}{  0,   0,   0}
\definecolor{cb-blue-green} {RGB}{  0,  073,  073}
\definecolor{cb-green-sea}  {RGB}{  0, 146, 146}
\definecolor{cb-rose}       {RGB}{255, 109, 182}
\definecolor{cb-salmon-pink}{RGB}{255, 182, 119}
\definecolor{cb-purple}     {RGB}{ 73,   0, 146}
\definecolor{cb-blue}       {RGB}{ 0, 109, 219}
\definecolor{cb-lilac}      {RGB}{182, 109, 255}
\definecolor{cb-blue-sky}   {RGB}{109, 182, 255}
\definecolor{cb-blue-light} {RGB}{182, 219, 255}
\definecolor{cb-burgundy}   {RGB}{146,   0,   0}
\definecolor{cb-brown}      {RGB}{146,  73,   0}
\definecolor{cb-clay}       {RGB}{219, 209,   0}
\definecolor{cb-green-lime} {RGB}{ 36, 255,  36}
\definecolor{cb-yellow}     {RGB}{255, 255, 109}

\usepackage{times}
\usepackage{latexsym}
\usepackage{amsfonts,amsmath}
\usepackage{array}
\usepackage{graphicx}
\usepackage{subcaption}
\usepackage{multirow}
\usepackage{changepage}
\usepackage{booktabs}

\usepackage{algorithm2e}
\usepackage{setspace}
\usepackage{breqn}
\usepackage{bbm}
\usepackage{dirtytalk}
\usepackage{graphicx}
\usepackage{float}
\usepackage{listings}
\usepackage{booktabs}
\usepackage{caption}
\usepackage{enumitem}
\usepackage{amsmath,stackengine}
\usepackage{cleveref}
\usepackage{float}
\usepackage[T1]{fontenc}

\usepackage[utf8]{inputenc}

\usepackage{microtype}

\graphicspath{ {./images/} }
\title{Shaking Syntactic Trees on the Sesame Street: Multilingual Probing with Controllable Perturbations}

\author{Ekaterina Taktasheva\thanks{$^*$ \texttt{etaktasheva@hse.ru}} \\
  HSE University \\
  Moscow, Russia \\
  \And
  Vladislav Mikhailov \\
  SberDevices, Sberbank \\
  Moscow, Russia \\
  \And
  Ekaterina Artemova \\
  HSE University \\
  Huawei Noah’s Ark lab \\
  Moscow, Russia
  }

\begin{document}
\maketitle
\begin{abstract}
Recent research has adopted a new experimental field centered around the concept of text perturbations which has revealed that shuffled word order has little to no impact on the downstream performance of Transformer-based language models across many NLP tasks. These findings contradict the common understanding of how the models encode hierarchical and structural information and even question if the word order is modeled with position embeddings. To this end, this paper proposes nine probing datasets organized by the type of \emph{controllable} text perturbation for three Indo-European languages with a varying degree of word order flexibility: English, Swedish and Russian. Based on the probing analysis of the M-BERT and M-BART models, we report that the syntactic sensitivity depends on the language and model pre-training objectives. We also find that the sensitivity grows across layers together with the increase of the perturbation granularity. Last but not least, we show that the models barely use the positional information to induce syntactic trees from their intermediate self-attention and contextualized representations.
\end{abstract}

\section{Introduction}
An extensive body of works is devoted to analyzing syntactic knowledge of Transformer language models (LMs) \cite{vaswani2017attention,clark-etal-2019-bert,goldberg2019assessing,belinkov-glass-2019-analysis}. BERT-based LMs \cite{devlin-etal-2019-bert} have demonstrated their abilities to encode various linguistic and hierarchical properties \cite{lin-etal-2019-open,jawahar2019does,jo2020roles} which have a positive effect on the downstream performance \cite{liu-etal-2019-linguistic,miaschi-etal-2020-linguistic} and serve as an inspiration for syntax-oriented architecture improvements \cite{wang2019structbert,bai-etal-2021-syntax,ahmad-etal-2021-syntax,sachan-etal-2021-syntax}. Besides, a variety of pre-training objectives has been introduced \cite{liu2020survey}, with some of them modeling reconstruction of the perturbed word order \cite{lewis-etal-2020-bart,tao-etal-2021-learning,panda-etal-2021-shuffled}.

Recent research has adopted a new experimental direction aimed at exploring the syntactic knowledge of LMs and their sensitivity to word order employing \emph{text perturbations} \cite{futrell2018rnns,futrell-etal-2019-neural,ettinger-2020-bert}. Some studies show that shuffling word order causes significant performance drops on a wide range of QA tasks \cite{si2019does,sugawara2020assessing}. However, a number of works demonstrates that such permutation has little to no impact during the pre-training and fine-tuning stages \cite{pham2020out,sinha2020unnatural,DBLP:journals/corr/abs-2104-06644,o2021context,hessel-schofield-2021-effective,gupta2021bert}. The latter contradict the common understanding on how the hierarchical and structural information is encoded in LMs \cite{rogers-etal-2020-primer}, and even may question if the word order is modeled with the position embeddings \cite{wang2020position,dufter2021position}.

This has stimulated a targeted probing of the LMs internal representations generated from original texts and their permuted counterparts \cite{DBLP:journals/corr/abs-2104-06644,hessel-schofield-2021-effective}. A new type of \emph{controllable} probes has been proposed, designed to test the LMs sensitivity to granular character- and sub-word level manipulations \cite{clouatre2021demystifying}, as well as structured syntactic perturbations \cite{alleman-etal-2021-syntactic}. Despite the emerging interest in the field, little is investigated for languages other than English, specifically those with flexible word order. 

This paper extends the ongoing research on the syntactic sensitivity to three Indo-European languages with a varying degree of word order flexibility: English, Swedish, and Russian. The contributions of this work are summarized as follows. First, we propose nine probing datasets in the languages mentioned above, organized by the type of controllable syntactic perturbation: N-gram perturbation (\textbf{NgramShift}), shuffling parts of the syntactic clauses (\textbf{ClauseShift}) and randomizing word order (\textbf{RandomShift}). Despite that randomizing word order has been studied from many perspectives (see Section \ref{related_work}), \textbf{NgramShift} differs from similar approaches \cite{conneau-etal-2018-cram,ravishankar-etal-2019-probing,eger-etal-2020-probe,alleman-etal-2021-syntactic} in that the N-grams correspond to \emph{only} syntactic phrases (e.g. prepositional or numerical phrases) rather than random word spans. \textbf{ClauseShift} is a previously unexplored type of syntactic perturbation adopted from the syntactic tree augmentation method \cite{sahin-steedman-2018-data}. Second, we apply a combination of parameter-free interpretation methods to test the sensitivity of two multilingual Transformer LMs: M-BERT \cite{devlin-etal-2019-bert}, and M-BART \cite{liu-etal-2020-multilingual-denoising}. We hypothesize that M-BART is more robust to the perturbations as opposed to M-BERT since it is learned to restore the shuffled input during pre-training. We evaluate the discrepancy in the syntactic trees induced by the models from perturbed sentences against the original ones, along with the ability to distinguish between them by judging their linguistic acceptability \cite{lau-etal-2020-furiously}. Finally, we analyze the relationship between the models' probe performance and position embeddings (PEs). To the best of our knowledge, it is one of the first attempts to introspect PEs regarding structural probing, particularly in the light of syntactic perturbations. The code and datasets are publicly available\footnote{\url{https://github.com/evtaktasheva/dependency_extraction}}.

\section{Related Work} 
\label{related_work}
\paragraph{Syntax Probing}
Most of the previous studies on the syntactic knowledge of LMs are centered around the concept of probing tasks, where a simple classifier is trained to predict a particular linguistic property based on the model internal representations \cite{conneau-etal-2018-cram}. The scope of the properties ranges from dependency relations \cite{tenney2018you} to the depth of a syntax tree, and top constituents \cite{conneau-etal-2018-cram}. A variety of probing datasets and benchmarks have been developed. To name a few, \citet{liu-etal-2019-linguistic} create a probing suite focused on fine-grained linguistic phenomena, including hierarchical knowledge. SyntaxGym \cite{gauthier2020syntaxgym} and LINSPECTOR \cite{sahin-etal-2020-linspector} allow for targeted evaluation of the LMs linguistic knowledge in a standardized and reproducible environment. 

These studies have proved that LMs are capable of encoding linguistic and hierarchical information \cite{belinkov-glass-2019-analysis,rogers-etal-2020-primer}. However, the probing paradigm has been lately criticized for relying on \emph{supervised} probes, which can learn linguistic properties given the supervision, and make it challenging to interpret the results because of the additional set of parameters \cite{hewitt-liang-2019-designing,belinkov2021probing}. Towards that end, \citet{hewitt-manning-2019-structural} introduce a \emph{structural} probe to explore a linear transformation of the embedding space, which best approximates the distance between words and depth of the parse tree. The method has proved to infer the hierarchical structure without any linguistic annotation \cite{kim2020pre}. \citet{maudslay2021syntactic} propose a \emph{Jabberwocky} probing suite of semantically nonsensical but syntactically well-formed sentences. The results demonstrate that the BERT-based LMs do not isolate semantics from syntax, which motivates further development of the probing field.

\paragraph{Acceptability Judgements} Another line of works relies on the concept of acceptability judgments. The CoLA benchmark \cite{warstadt-etal-2019-neural} and its counterpart for Swedish \cite{volodina2021dalaj} test LMs ability to identify various linguistic violations. Although Transformer LMs have outperformed the CoLA human solvers on the GLUE leaderboard \cite{wang-etal-2018-glue}, a granular linguistic analysis \cite{DBLP:journals/corr/abs-1901-03438} shows that the models struggle with long-distance syntactic phenomena as opposed to more local ones. Similar in spirit, BLiMP \cite{warstadt-etal-2020-blimp-benchmark}, and CLiMP \cite{xiang-etal-2021-climp} allow to evaluate the LMs with respect to the acceptability contrasts, framing the task as ranking sentences in minimal pairs. 

\paragraph{Text Perturbations} Recent research has adopted a scope of novel approaches to investigating the LMs sensitivity to syntax corruption and input data manipulations. Starting from studies on randomized word order in LSTMs \cite{hill-etal-2016-learning,khandelwal-etal-2018-sharp,sankar2019neural,nie2019analyzing}, text perturbations have emerged as an audacious experimental direction under the ``pre-train \& fine-tune'' paradigm along with the interpretation methods of modern LMs. \citet{si2019does,sugawara2020assessing} show that N-gram permutations and shuffled word order in the fine-tuning data cause BERT's performance drops up to 22\% on a wide range of QA tasks. In contrast, several works report that models fine-tuned on such perturbed data still produce high confidence predictions and perform close to their counterparts on many tasks, including the GLUE benchmark \cite{ahmad-etal-2019-difficulties,sinha2020unnatural,liu2021importance,hessel-schofield-2021-effective,gupta2021bert}. Similar results are demonstrated by the RoBERTa model \cite{liu2019roberta} when the word order perturbations are incorporated into the pre-training objective \cite{panda-etal-2021-shuffled} or tested as a part of full pre-training on the perturbed corpora \cite{DBLP:journals/corr/abs-2104-06644}. \citet{DBLP:journals/corr/abs-2104-06644} find that the randomized RoBERTa models are similar to their naturally pre-trained peer according to parametric probes but perform worse according to the non-parametric ones.

Recognizing the need to further explore the LMs sensitivity to word order, \citet{clouatre2021demystifying} and \citet{alleman-etal-2021-syntactic} conduct the interpretation analysis of LMs by means of \emph{controllable} text perturbations. \citet{clouatre2021demystifying} propose two metrics that score local and global structure of sentences perturbed at the granularity of characters and sub-words. The metrics allow identifying that both conventional and Transformer LMs rely on the local order of tokens more than the global one. \citet{alleman-etal-2021-syntactic} find that BERT builds syntactic complexity towards the output layer and demonstrates a growing sensitivity to the hierarchical phrase structure across layers. In line with these studies, we analyze the syntactic sensitivity of Transformer-based LMs, extending the experimental setup to the multilingual setting.

\section{Controllable Perturbations}
This work proposes three types of \emph{controllable} syntactic perturbations varying in the extent of sentence corruption. We construct nine probing tasks\footnote{We use sentences from the CoNLL 2017 Shared Task on Multilingual Parsing from Raw Texts to Universal Dependencies \cite{ginter@conll}.} for three Indo-European languages\footnote{\url{https://wals.info}}: English (West Germanic, analytic), Swedish (North Germanic, analytic), and Russian (Balto-Slavic, fusional). Based on the dominant constituent order, all three languages are classified as the SVO (Subject-Verb-Object) languages. Nevertheless, there are some differences between them regarding word order flexibility. Russian is known to exhibit free word order as all of the possible constituent reorderings are acceptable: SOV, OSV, SVO, OVS, VSO, VOS \cite{bailyn2012syntax}. English allows for only two of them, namely SVO and OSV \cite{prince1988pragmatic}. Swedish belongs to the verb-second languages, which poses different restrictions on the possible constituent reorderings \cite{borjars2003subject}. Each dataset\footnote{A brief statistics is outlined in Appendix~\ref{app:stata}.} consists of 10k pairs of the corresponding perturbed sentence and its original. 

\vspace{0.5em}\noindent \textbf{NgramShift} tests the LM sensitivity to \emph{local} perturbations taking into account the syntactic structure. We used a set of carefully designed morphosyntactic patterns to perturb N-grams that correspond to \emph{only} syntactic phrases such as numeral phrases, determiner phrases, compound noun phrases, prepositional phrases, etc. Towards this, we applied TF-IDF weighting from scikit-learn library \cite{pedregosa2011scikit} to build a ranked N-gram feature matrix from the corpora and further used it for the N-gram inversion. We used the N-gram range $\in [2; 4]$ for each language. Note that the number of words that change their absolute positions is similar for different values of $N$. Figure \ref{fig:example-nshift} illustrates the shift of the head in the prepositional phrase \textit{``to school''} for the sentence \textit{``He did not go to school yesterday''}.

\begin{figure}[ht]
    \centering
    \fbox{\begin{minipage}{0.95\columnwidth}
    \setstretch{1.5}
    \parbox{\columnwidth}{
    	    \centering
            
            \textcolor{cb-blue-green}{He did not go} \textcolor{cb-brown}{to} \textcolor{cb-blue}{school} \textcolor{cb-rose}{yesterday} \\
            
            \textbf{En:} \textcolor{cb-blue-green}{He did not go} \textcolor{cb-blue}{school} \textcolor{cb-brown}{to} \textcolor{cb-rose}{yesterday}\\
            
            \textbf{Ru:} \textcolor{cb-rose}{Vchera} \textcolor{cb-blue-green}{on ne poshel} \textcolor{cb-blue}{shkolu} \textcolor{cb-brown}{v} \\
            
            \textbf{Sv:} \textcolor{cb-blue-green}{Han gick inte} \textcolor{cb-blue}{skolan} \textcolor{cb-brown}{till} \textcolor{cb-rose}{igår}
        }
     \end{minipage}
     }
    \caption{Examples of the N-gram perturbations (\textbf{NgramShift}). Languages:  \textbf{En}=English, \textbf{Ru}=Russian, \textbf{Sv}=Swedish. The English sentence is translated to the other languages for illustrational purposes.
    }
    \label{fig:example-nshift}
\end{figure}

\vspace{0.5em}\noindent \textbf{ClauseShift} probes the LM sensitivity to \emph{distant} perturbations at the level of syntactic clauses. We use the syntactic tree augmentation method \cite{csahin2019data} to rotate sub-trees around the root of the dependency tree of each sentence to form a new synthetic sentence. We then apply a set of manually curated language-specific heuristics to filter out sentences uncorrupted by the rotation procedure. Figure \ref{fig:example-clauseshift} outlines an example of the clause rotation perturbation for the sentence \textit{``He manages to tell her that she has been resurrected''}.

\begin{figure}[ht]
    \centering
    \fbox{\begin{minipage}{0.95\columnwidth}
    \setstretch{1.5}
    \parbox{\columnwidth}{
    	    \centering
            \small
            \textcolor{cb-blue-green}{He manages to tell her} \textcolor{cb-rose}{that she has been resurrected}\\
            
            \textbf{En:} \textcolor{cb-rose}{That she has been resurrected} \textcolor{cb-blue-green}{he manages to tell her} \\
            
            \textbf{Sv:} \textcolor{cb-rose}{Att hon har uppstått} \textcolor{cb-blue-green}{han lyckas berätta för henne}\\
            
            \textbf{Ru:} \textcolor{cb-rose}{Chto ona byla voskreshena} \textcolor{cb-blue-green}{on smog rasskazat' ej}
            }
     \end{minipage}
     }
    \caption{Examples of the clause rotation perturbation (\textbf{ClauseShift}). Languages:  \textbf{En}=English, \textbf{Ru}=Russian, \textbf{Sv}=Swedish. The English sentence is translated to the other languages for illustrational purposes.
    }
    \label{fig:example-clauseshift}
\end{figure}

\vspace{0.5em}\noindent \textbf{RandomShift} tests the LM sensitivity to \emph{global} perturbations obtained by shuffling the word order. This type represents an extreme case of sentence permutation and is useful for comparing the behavior of the models at the scale of the perturbation complexity. An example of the randomized word order perturbation for the sentence \textit{``She wanted to go to London''} is presented in Figure \ref{fig:example-randomshift}.

\begin{figure}[ht]
    \centering
    \fbox{\begin{minipage}{0.95\columnwidth}
    \setstretch{1.5}
    \parbox{\columnwidth}{
    \centering
    \textcolor{cb-blue-green}{She} \textcolor{cb-brown}{wanted} 
    \textcolor{cb-blue}{to go} 
    \textcolor{cb-rose}{to London}\\

    \textbf{En:} \textcolor{cb-brown}{Wanted} \textcolor{cb-rose}{London} 
    \textcolor{cb-blue}{go} 
    \textcolor{cb-blue-green}{she} 
    \textcolor{cb-blue}{to} 
    \textcolor{cb-rose}{to} \\
    
    \textbf{Sv:} \textcolor{cb-brown}{Ville} \textcolor{cb-rose}{London} 
    \textcolor{cb-blue}{åka} 
    \textcolor{cb-blue-green}{hon} 
    \textcolor{cb-rose}{till} 
    \textcolor{cb-blue}{att}\\
    
    \textbf{Ru:} \textcolor{cb-brown}{Hotela} \textcolor{cb-rose}{London} 
    \textcolor{cb-blue}{poehat'} 
    \textcolor{cb-blue-green}{ona} 
    \textcolor{cb-rose}{v}}
     \end{minipage}
     }
    \caption{Examples of the word order shuffling (\textbf{RandomShift}). Languages: \textbf{En}=English, \textbf{Ru}=Russian, \textbf{Sv}=Swedish. The English sentence is translated to the other languages for illustrational purposes.
    }
    \label{fig:example-randomshift}
\end{figure}

\section{Experimental Setup} 
\label{setup}

\subsection{Models}
\label{setup:models}
The experiments are run on two 12-layer multilingual Transformer models released by the HuggingFace library \cite{wolf-etal-2020-transformers}:

\vspace{0.5em}\noindent \textbf{M-BERT}\footnote{Model name: \texttt{bert-base-multilingual-cased}.} is pre-trained using masked language modeling (MLM) and next sentence
prediction objectives, over concatenated monolingual
Wikipedia corpora in 104 languages.

\vspace{0.5em}\noindent \textbf{M-BART}\footnote{Model name: \texttt{facebook/mbart-large-cc25}.} is a sequence-to-sequence model that comprises a BERT encoder and an autoregressive GPT-2 decoder \cite{radford2019language}. The model is pre-trained on the CC25 corpus in 25 languages using text infilling and sentence shuffling objectives, where it learns to predict masked word spans and reconstruct the permuted input. We use only the encoder in our experiments.

\subsection{Interpretation Methods}
\label{setup:methods}

\paragraph{Parameter-free Probing} We apply two unsupervised probing methods to reconstruct syntactic trees from self-attention (\textbf{Self-Attention Probing}) and so-called ``impact'' (\textbf{Token Perturbed Masking}) matrices computed by feeding the MLM models with each sentence $s$ and its perturbed version $s'$. The trees are induced by  Chu-Liu-Edmonds algorithm \cite{chu1965shortest,edmonds1968optimum} used to compute the Maximum Spanning Tree starting from the root of the corresponding gold dependency tree \cite{raganato-tiedemann-2018-analysis,htut2019attention,wu-etal-2020-perturbed}. The probing performance is evaluated by the Undirected Unlabeled Attachment Score (UUAS), which reflects the percentage of words that have been assigned the correct head without taking the direction of relations and dependency labels into account \cite{klein-manning-2004-corpus}. 

\vspace{0.5em}\noindent \textbf{Self-Attention Probing} \cite{htut2019attention} allows to explore if attention heads encode complete syntactic trees. To this end, each layer-head attention matrix is treated as a weighted directed graph where the vertices represent words in the input sentence and edges are the attention weights. Model-specific special tokens such as \texttt{[CLS]}, \texttt{[SEP]}, \texttt{<s>}, \texttt{</s>} are excluded at the pre-processing stage to eliminate their impact on other tokens.

\vspace{0.5em}\noindent \textbf{Token Perturbed Masking} \cite{wu-etal-2020-perturbed} extracts global syntactic information by measuring the impact one word has on the prediction of another in an MLM. The impact matrix is similar to the self-attention matrix as it reflects the inter-word relationships in terms of Euclidean distance, except that it is derived from the outputs of the MLM head. For the sake of space, we refer the reader to \citet{wu-etal-2020-perturbed} for more details.

\paragraph{Representation Analysis} \citet{hessel-schofield-2021-effective} propose two metrics to compare contextualized representations and self-attention matrices produced by the model for each pair of sentences $s$ and $s'$. \textit{Token Identifiability (TI)} evaluates the similarity of the LM's contextualized representations of a particular token in $s$ and $s'$. It is high if the token representations are similar to one another. \textit{Self-Attention Distance (SAD)} measures if each token in $s$ relates to similar words in $s'$ by computing row-wise Jensen-Shannon Divergence between the two self-attention matrices. It is low if an LM attends to the same words despite the perturbations.

\paragraph{Pseudo-perplexity} Pseudo-perplexity (PPPL) is an intrinsic measure that estimates the probability of a sentence with an MLM similar to that of conventional LMs \cite{salazar-etal-2020-masked}. PPPL-based measures have proved to correlate with human ratings \cite{lau2017grammaticality}, match or outperform autoregressive LMs (GPT-2) in ranking hypotheses for downstream tasks and the BLiMP benchmark \cite{salazar-etal-2020-masked}, and perform at the human level in acceptability judgments \cite{lau-etal-2020-furiously}. We use two PPPL-based measures under implementation\footnote{\url{https://github.com/jhlau/acceptability-prediction-in-context}} by \citet{lau-etal-2020-furiously} to infer probabilities of the sentences and their perturbed counterparts. The \textit{MeanLP} and \textit{PenLP} measures are computed as the sum of pseudo-log-likelihood scores for each token in the sentence normalized by the total number of tokens. \textit{PenLP} additionally scales the denominator with the exponent $\alpha$ to penalize the effect of high scores.

\subsection{Positional Encoding} Various PEs have been proposed to utilize the information about word order in the Transformer-based LMs \cite{wang2020position,dufter2021position}. Surprisingly, little is known about what PEs capture and how well they learn the meaning of positions. \citet{wang-chen-2020-position} among the first present an extensive study on the properties captured by PEs in different pre-trained Transformers and empirically evaluate their impact on the downstream performance for many NLP tasks. In the spirit of this work, we aim at analyzing the impact of the PEs on the syntactic probe performance. Towards this end, we consider the following three configurations of PEs of the M-BERT and M-BART models: (1) \textbf{absolute}=frozen PEs; (2) \textbf{random}=randomly initialized PEs; and (3) \textbf{zero}=zeroed PEs.

\section{Results} 
\label{res}

\begin{figure*}[t!]
  \centering
  \includegraphics[width=.7\textwidth]{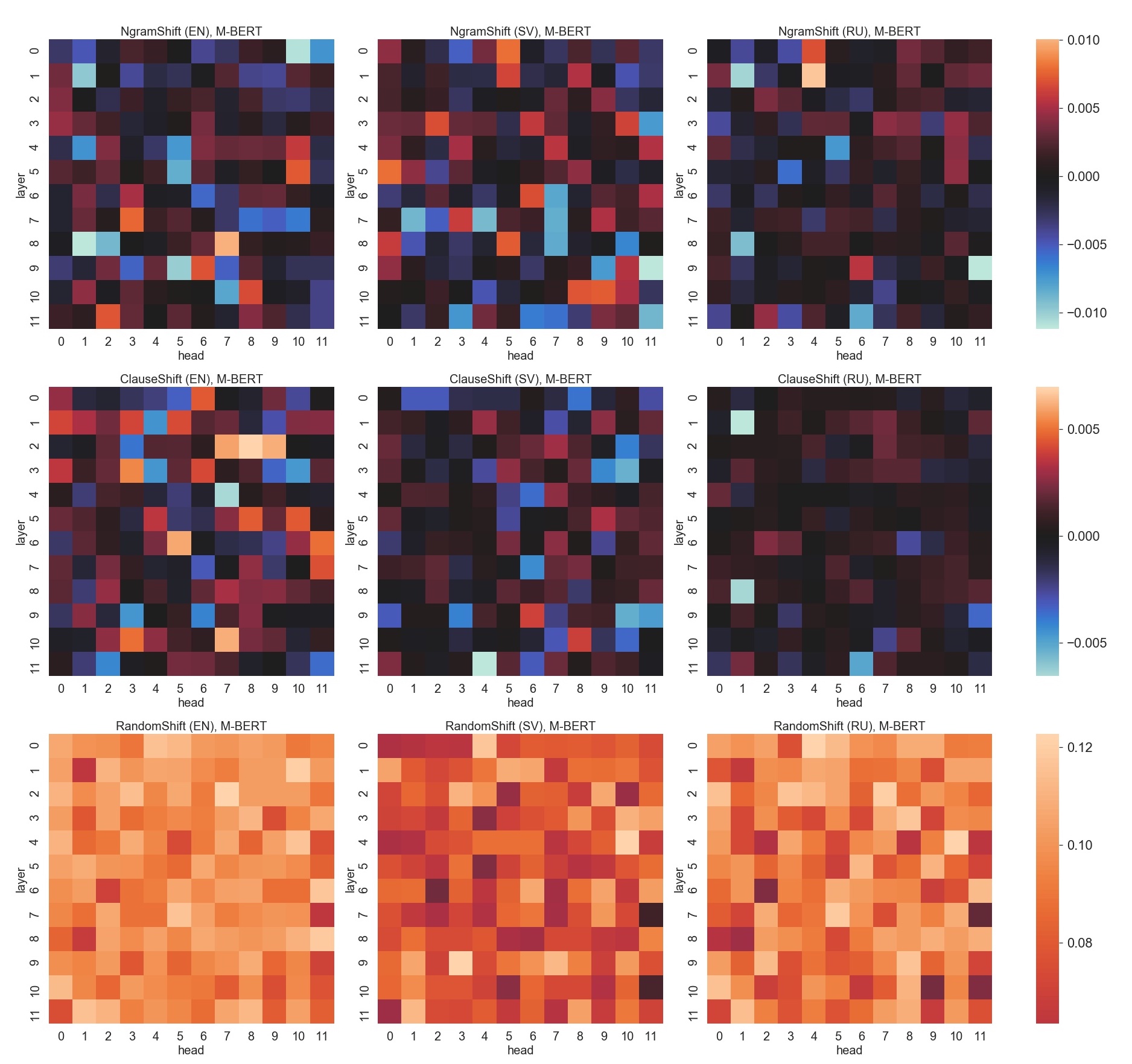}
  \caption{The task-wise heatmaps depicting the $\delta$ UUAS scores by M-BERT for each language. Method=\textbf{Self-Attention Probing}. PE=\textbf{absolute}. X-axis=Attention head index. Y-axis=Layer index. Tasks: \textbf{NgramShift} (top); \textbf{ClauseShift} (middle); \textbf{RandomShift} (bottom). Languages: \textbf{En}=English (left); \textbf{Sv}=Swedish (middle); \textbf{Ru}=Russian (right).}
  \label{fig:mbert_complexity}
\end{figure*}

\subsection{Parameter-free Probing}
\label{pfp}
The discrepancy in the syntactic trees induced from the original sentences and their perturbed analogs is measured as the difference between the corresponding UUAS scores ($\delta$ UUAS). The lower the $\delta$ UUAS, the better is the syntax tree reconstructed from $s'$ with respect to the UUAS score for $s$. 

\begin{figure*}[h]
    \centering
    \begin{subfigure}[b]{0.6\linewidth}
    \centering
    \includegraphics[width=\linewidth]{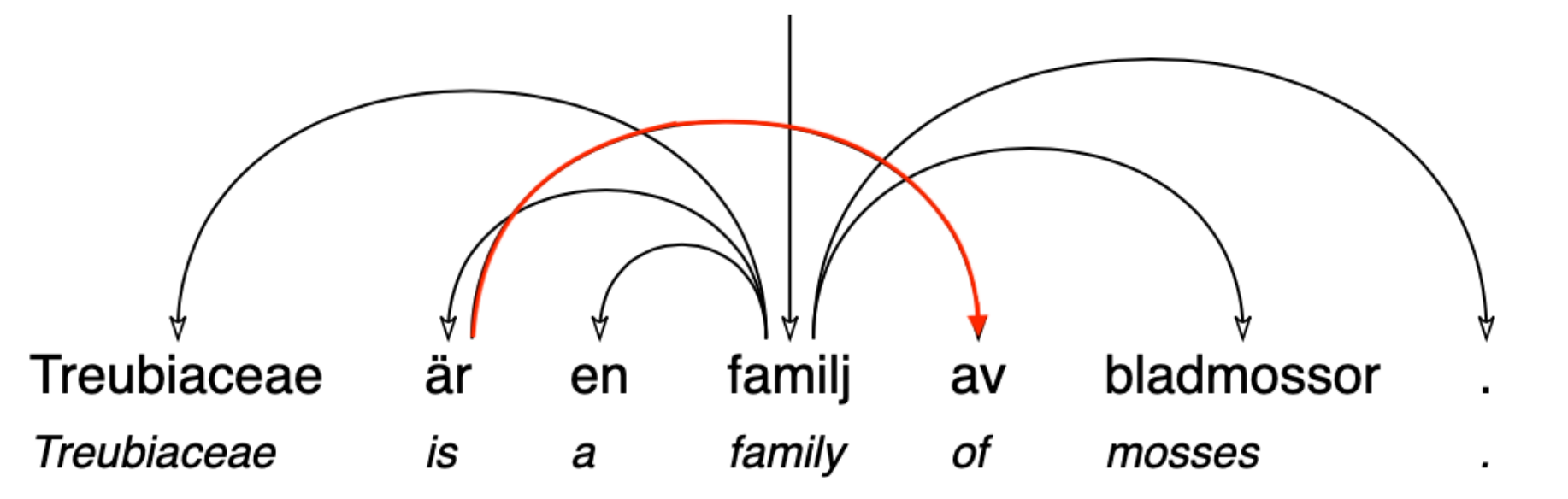}
    \caption{\texttt{original}}
    \label{fig:tr-sv-nshift-gr}
    \end{subfigure}
    \begin{subfigure}[b]{0.6\linewidth}
    \centering
    \includegraphics[width=\linewidth]{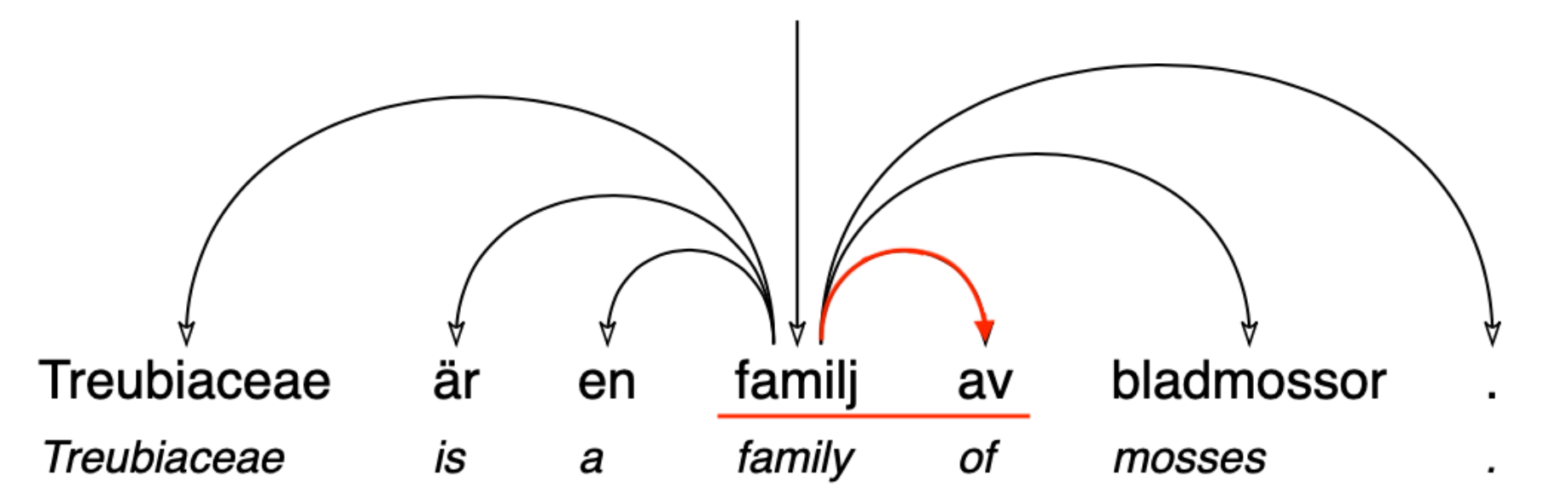}
    \caption{\texttt{perturbed}}
    \label{fig:tr-sv-nshift-ungr}
    \end{subfigure}  
    \begin{subfigure}[b]{0.6\linewidth}
    \centering
    \includegraphics[width=\linewidth]{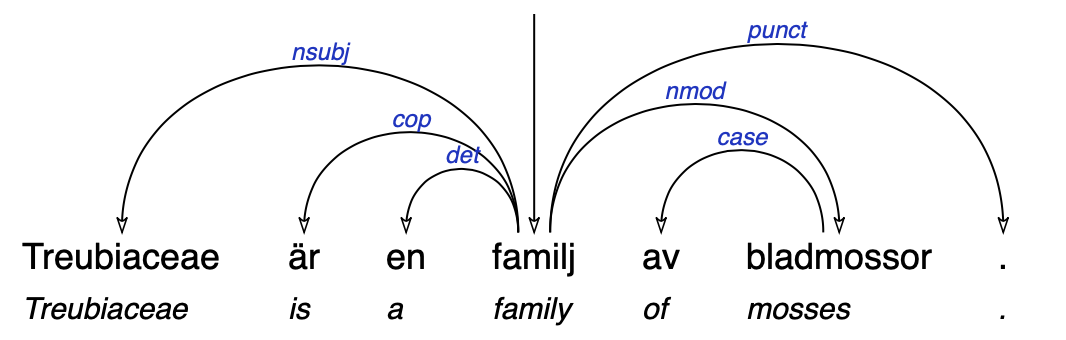}
    \caption{\texttt{gold}}
    \label{fig:tr-sv-nshift-ud}
    \end{subfigure}
    \caption{Graphical representations of the syntactic trees inferred for the Swedish sentence \textit{Treubiaceae är en familj av bladmossor} 'Treubiaceae is a family of mosses' and its perturbed version. \texttt{original}=the original sentence; \texttt{perturbed}=the perturbed version; \texttt{gold}=gold standard. Task=\textbf{NgramShift}. Model=\textbf{M-BERT} (Layer: 11; Head: 2). Method=\textbf{Self-Attention Probing}. The perturbation is underlined with red, and incorrectly assigned dependency heads are marked with red arrows.
    }
    \label{fig:tr-sv}
\end{figure*}

\paragraph{Self-Attention Probing} Figures \ref{fig:mbert_complexity} and \ref{fig:mbart_complexity} in Appendix \ref{app:pfp} outline the task-wise heatmaps with the $\delta$ UUAS scores achieved by the M-BERT and M-BART models with \textbf{absolute} PEs for each layer-head pair, respectively. The models exhibit similar behavior, demonstrating positive correlation between the $\delta$ UUAS scores and the granularity of the perturbation. The overall pattern for both models is that they display little to no sensitivity to \emph{local} and \emph{distant} perturbations (\textbf{NgramShift}, \textbf{ClauseShift}) in contrast to the \emph{global} ones (\textbf{RandomShift}). We provide examples of the dependency trees extracted from the self-attention matrices of the M-BERT model for the Swedish \textbf{NgramShift} task on Figure \ref{fig:tr-sv}. The trees from both original (see Figure \ref{fig:tr-sv-nshift-gr}) and perturbed (see Figure \ref{fig:tr-sv-nshift-ungr}) sentence versions receive the UUAS score of 0.86, demonstrating little changes in the assigned dependency heads under the local perturbation. On the contrary, randomizing word order (\textbf{RandomShift}) corrupts the syntactic structure significantly with a $\delta$ UUAS score of 0.33 (see Figure \ref{fig:tr-en}, Appendix \ref{app:pfp}).

\begin{figure*}[h!]
  \centering
  \includegraphics[width=.82\textwidth]{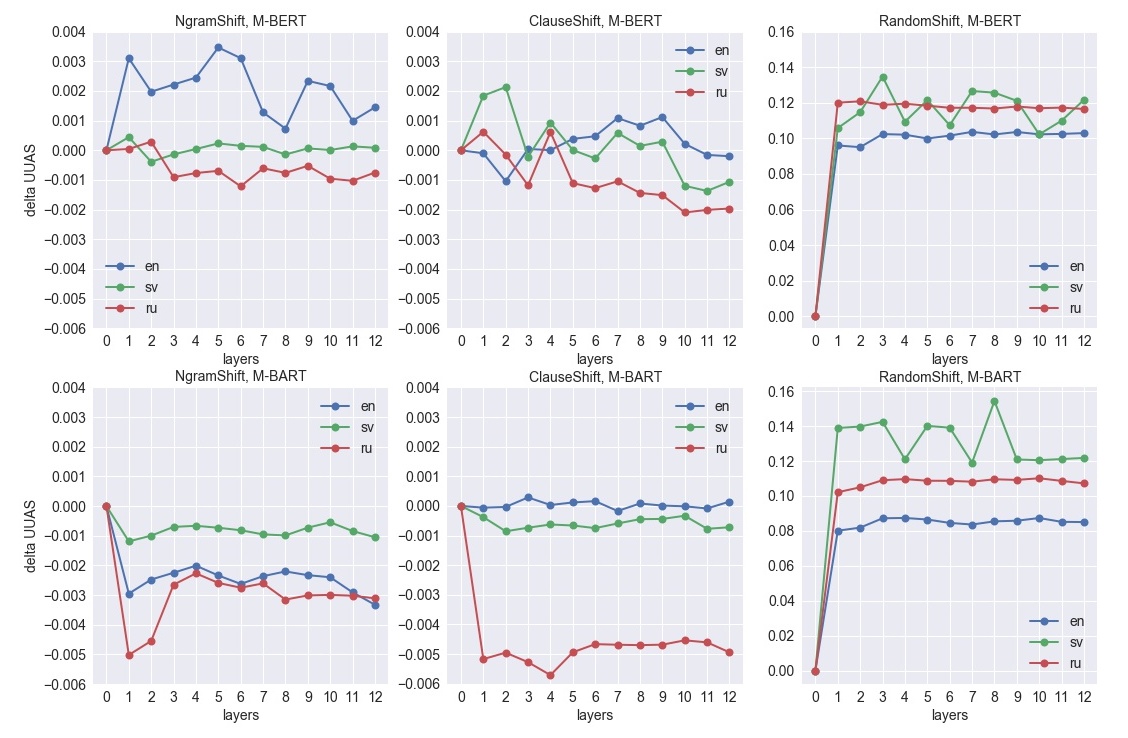}
  \caption{The probing performance in $\delta$ UUAS  across layers under \textbf{Token Perturbed Probing}. PE=\textbf{absolute}. The scores are averaged over attention heads at each layer. X-axis=Attention head index. Y-axis=$\delta$ UUAS.}
  \label{fig:perturbed}
\end{figure*}

\paragraph{Token Perturbed Masking} The models show similar results to that of in \textbf{Self-Attention Probing}, with regards to the perturbation granularity (see Figure \ref{fig:perturbed}). In spite of that, the model performance on the \textbf{NgramShift} and \textbf{ClauseShift} reveal some differences between the encoders. M-BART generally achieves lower and close to zero $\delta$ UUAS scores, meaning to better restore the hierarchical information from the perturbed sentences (e.g., \textbf{ClauseShift}: [\textbf{Sv, Ru}]). We relate this to the fact that M-BART is pre-trained with the sentence shuffling objective.

\paragraph{Language-wise Comparison} Another observation is that there are more insensitive attention heads on the Russian tasks, possibly indicating that it is harder to distinguish from the perturbations as opposed to English and Swedish, particularly on the \textbf{ClauseShift} task with typically longer and syntactically more complex sentences (see Figures \ref{fig:mbert_complexity}, \ref{fig:mbart_complexity}, Appendix \ref{app:pfp}). As for Swedish, which has a similar to English but stricter syntactic structure, M-BART tends to induce correct syntactic trees from the permuted sentences more frequently. This is indicated by negative $\delta$ UUAS scores on most tasks.

\paragraph{Positional Encoding} Analysis of the positional encoding shows that despite the genuine belief that positional information contributes most to syntactic structure encoding, the models do not seem to rely on it as much as might be expected. Figure \ref{fig:mbert_complexity_pe} (see Appendix \ref{app:pfp}) illustrates the distribution of $\delta$ UUAS scores for M-BERT with different PEs on English tasks. The heatmaps show that \textbf{zero} and \textbf{random} PEs only slightly affects the quality of the probe performance of the self-attention heads.

\begin{figure*}[t!]
  \centering
  \includegraphics[width=.82\textwidth]{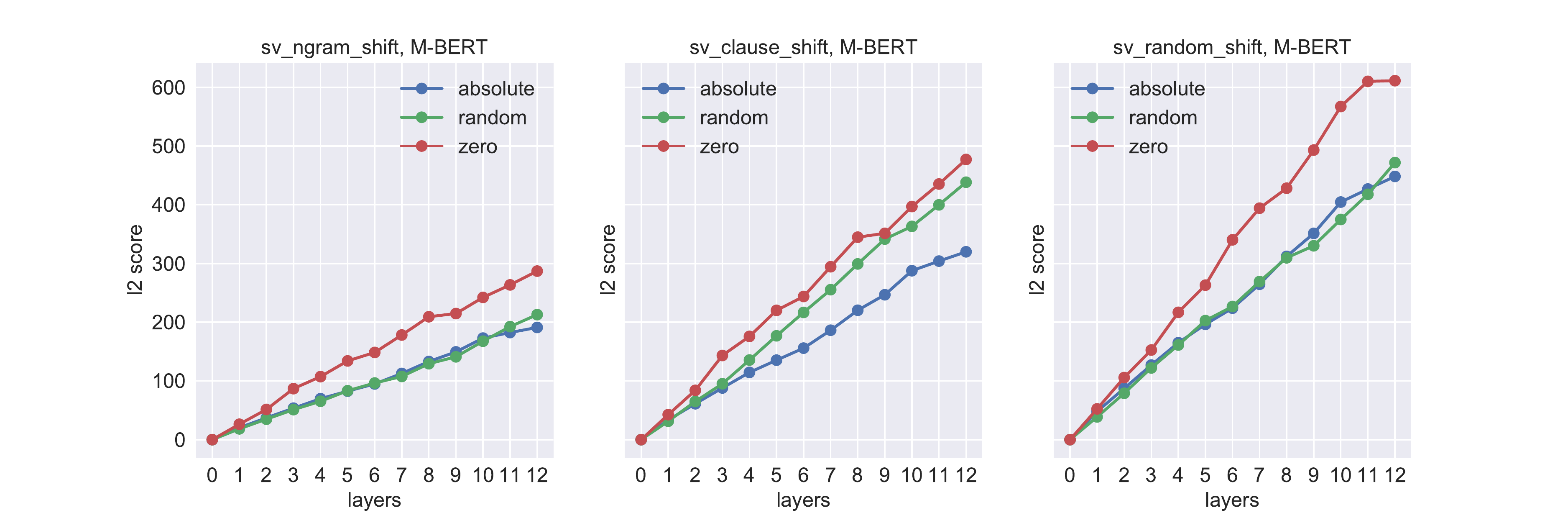}
  \caption{The Euclidean distances between the impact matrices computed by M-BERT with different PEs over each pair of sentences ($s$, $s'$) for Swedish. The distances are averaged over attention heads at each layer. Method: \textbf{Token Perturbed Masking}. Tasks: \textbf{NgramShift} (left); \textbf{ClauseShift} (middle); \textbf{RandomShift} (right)}. 
  \label{fig:l2-perturbed-sv-mbert}
\end{figure*}

To analyze the impact of PEs from another perspective, for each pair of ($s$, $s'$) we compute the Euclidean distance (L2) between the corresponding impact (\textbf{Token Perturbed Probing}) and self-attention matrices (\textbf{Self-Attention Probing}) described in Section \ref{setup:methods}. The difference in the impact matrices produced by M-BERT model is generally observed only in the setting with \textbf{zero} PEs (see Figures \ref{fig:l2-perturbed-sv-mbert}; Figures \ref{fig:l2-perturbed-ru-mbert}-\ref{fig:l2-perturbed-en-mbert}, Appendix \ref{app:pfp}). In contrast, there is almost no difference between the representations generated by M-BART across all configurations of the PEs (see Figures \ref{fig:l2-perturbed-sv-mbart}-\ref{fig:l2-perturbed-ru-mbart}, Appendix \ref{app:pfp}). This behavior is consistent with the head-wise results under \textbf{Self-Attention Probing} for all languages.

\subsection{Representation Analysis}
\label{res:repr}
\paragraph{Token Identifiability} The overall pattern for both models under the representation analysis is that for \emph{local} and \emph{distant} perturbations TI steadily decreases towards the output layer with rapid increases at layers $[1, 10]$ (see Figure \ref{fig:ti-ru}, Appendix \ref{app:repr}), and high for \emph{global} perturbations (\textbf{RandomShift}). TI decreases when the perturbed inputs generate embeddings different from the intact ones. Despite that higher layers in both models are more sensitive, the perturbed representations remain similar to that of the original \cite{hessel-schofield-2021-effective}.

\paragraph{Self-Attention Distance} The results by SAD show that both models score significantly lower with \textbf{random} and \textbf{zero} PEs (see Figure \ref{fig:sad-sv}, Appendix \ref{app:repr}), meaning lower sensitivity to the perturbations supported by the probing results (Section \ref{pfp}). This provides evidence that the encoders marginally rely on the positional information to induce the syntactic structure despite the distributions of the self-attention weights for the intact and perturbed sentences may differ according to the Jensen-Shannon divergence.

\subsection{Pseudo-perplexity} 
\label{res:pppl}
Consistent with the results under parameter-free probing (Section \ref{pfp}) and representation analysis (Section \ref{res:repr}), PPPL-based acceptability judgements\footnote{We present the results obtained by the \textit{MeanLP} measure which are consistent with those of \textit{PenLP}.} indicate that the encoders distinguish between the perturbations depending on their granularity. The overall trend is that for all languages the sentence pseudo-log-probability inferred from both LMs decreases with the increase of the perturbation complexity which is demonstrated by higher acceptability scores on \textbf{NgramShift}, but significantly lower scores on the \textbf{ClauseShift} and \textbf{RandomShift} (see Figures \ref{fig:pppl-mbart}-\ref{fig:pppl-mbert}, Appendix \ref{app:pppl}). The statistical significance of the PPPL distributions is confirmed with Kolmogorov–Smirnov and Wilcoxon signed-rank tests (p-value $<$ 0.01).

\section{Discussion}
\paragraph{The syntactic sensitivity depends upon language} At present, English remains the focal point of prior research in the field of NLP, leaving other languages understudied. Our probing experiments on the less explored languages with different word order flexibility show that M-BERT and M-BART behave slightly differently in Swedish and Russian. While M-BART better restores the corrupted syntactic structure on most of the tasks for Swedish, there are fewer attention heads sensitive to the perturbations in Russian, which is revealed through the examination of head-wise attention patterns of both models. Besides, the encoders receive lower probing performance for Russian that can be contributed to the more complex syntax and flexible word order.

\paragraph{Pre-training objectives can help to improve syntactic robustness} Analysis of the M-BERT and M-BART LMs that differ in the pre-training objectives shows that M-BERT achieves higher $\delta$ UUAS performance across all languages as opposed to M-BART pre-trained with the sentence shuffling objective. The lower $\delta$ UUAS probing performance indicates that M-BART better induces syntactic trees from both perturbed and intact sentences (see Section \ref{pfp}). Despite this, the representation and acceptability analysis demonstrate that M-BART is also capable of distinguishing between the perturbations (see Sections \ref{res:repr}-\ref{res:pppl}). A fruitful direction for future work is to analyze more LMs that differ in the architecture design and pre-training objectives.

\paragraph{The LMs are less sensitive to more granular perturbations} The results of the parameter-free probing show that M-BERT and M-BART exhibit little to no sensitivity to \emph{local} perturbations within syntactic groups (\textbf{NgramShift}) and \emph{distant} perturbations at the level of syntactic clauses (\textbf{ClauseShift}). In contrast, the \emph{global} perturbations (\textbf{RandomShift}) are best distinguished by the encoders. As the granularity of the syntactic corruption increases, we observe a worse probing performance under all considered interpretation methods. Namely, the results are supported by representation analysis metrics (see Section \ref{res:repr}) that indicate higher susceptibility to major changes in the sentences structure (\textbf{RandomShift, ClauseShift}), and the PPPL-based measures (see Section \ref{res:pppl}) prescribing higher acceptability scores to sentences with more granular perturbations (\textbf{NgramShift}). We also find that the sensitivity to the hierarchical corruption grows across layers together with the increase of the perturbation complexity, which is in line with \citet{alleman-etal-2021-syntactic}.

\paragraph{M-BERT and M-BART barely use positional information to induce syntactic trees} Previous research has shown that the token embeddings capture enough semantic information to restore the syntactic structure \cite{vilares2020parsing, kim2020pre, rosa2019inducing}. \citet{maudslay2021syntactic} claim that syntactic abilities of BERT-based LMs are overestimated and raise the problem of isolating semantics from syntax. However, more recent studies show that Transformer encoders encode redundant information \cite{luo2021positional}, may not sufficiently capture the meaning of positions and be unimportant for downstream tasks \cite{wang-chen-2020-position}, including the setting with perturbed fine-tuning data \cite{clouatre2021demystifying}. In spirit with the latter studies, our results under different PEs configurations reveal that M-BERT and M-BART do not need the precise position information to restore the syntactic tree from their internal representations. The overall behavior is that zeroed (except for M-BERT) or even randomly initialized PEs can result in the probing performance and one with absolute positions. We suppose that despite the absolute positions of words changes during the N-gram permutation and sub-tree rotation procedures, the word order within the clauses remains almost the same as in the intact sentence (\textbf{NgramShift}, \textbf{ClauseShift}). That is, the more granular perturbations marginally confuse the LMs when: (i) predicting the masked word under \textbf{Token Perturbation Probing} which can be performed using \emph{only} attention \cite{wang-chen-2020-position}, or (ii) judging the acceptability of the sentence where the low token pseudo-log-probability can occur at the juxtaposition of the syntactic groups, and clauses \cite{alleman-etal-2021-syntactic}. We leave a more detailed exploration of the relationship between PEs and probing analysis for future work.

\section{Conclusion}
This paper presents an extension of the ongoing research on the controllable text perturbations to the multilingual setting and introspection of positional embeddings in pre-trained LMs. We introduce nine probing datasets for three Indo-European languages varying in their flexibility of the word order: English, Swedish, and Russian. The suite is constructed using language-specific heuristics carefully designed under linguistic expertise and organized by three types of syntactic perturbations: randomization of word order studied by previous research from many perspectives and less explored permutations within syntactic phrases and clauses. The method includes a combination of parameter-free probing methods based on the intermediate self-attention and contextualized representations, novel metrics for representation analysis, and acceptability judgments with pseudo-perplexity. We conduct a line of experiments to probe the syntactic sensitivity of two multilingual Transformers, M-BERT and M-BART, the latter of which is learned to reconstruct the word order during pre-training. The LMs are less sensitive to more granular perturbations and build hierarchical complexity towards the output layer. The analysis of the understudied relationship between the position embeddings and syntactic probe performance reveals that the position information is not necessary for inducing the hierarchical structure, which is a promising direction for a more detailed investigation. The results also show that the syntactic sensitivity may depend on the language and be enhanced by pre-training objectives. We believe there is still room for exploring the sensitivity to word order and syntactic abilities of modern LMs, specifically across a more diverse set of languages and models varying in the architecture design choices.

\section*{Acknowledgements}

Ekaterina Taktasheva and Ekaterina Artemova are partially supported by the framework of the HSE University Basic Research Program.

\bibliography{anthology,custom}

\begin{thebibliography}{81}
\expandafter\ifx\csname natexlab\endcsname\relax\def\natexlab#1{#1}\fi

\bibitem[{Ahmad et~al.(2021)Ahmad, Li, Chang, and
  Mehdad}]{ahmad-etal-2021-syntax}
Wasi Ahmad, Haoran Li, Kai-Wei Chang, and Yashar Mehdad. 2021.
\newblock \href {https://doi.org/10.18653/v1/2021.acl-long.350}
  {Syntax-augmented multilingual {BERT} for cross-lingual transfer}.
\newblock In \emph{Proceedings of the 59th Annual Meeting of the Association
  for Computational Linguistics and the 11th International Joint Conference on
  Natural Language Processing (Volume 1: Long Papers)}, pages 4538--4554,
  Online. Association for Computational Linguistics.

\bibitem[{Ahmad et~al.(2019)Ahmad, Zhang, Ma, Hovy, Chang, and
  Peng}]{ahmad-etal-2019-difficulties}
Wasi Ahmad, Zhisong Zhang, Xuezhe Ma, Eduard Hovy, Kai-Wei Chang, and Nanyun
  Peng. 2019.
\newblock \href {https://doi.org/10.18653/v1/N19-1253} {On difficulties of
  cross-lingual transfer with order differences: A case study on dependency
  parsing}.
\newblock In \emph{Proceedings of the 2019 Conference of the North {A}merican
  Chapter of the Association for Computational Linguistics: Human Language
  Technologies, Volume 1 (Long and Short Papers)}, pages 2440--2452,
  Minneapolis, Minnesota. Association for Computational Linguistics.

\bibitem[{Alleman et~al.(2021)Alleman, Mamou, A~Del~Rio, Tang, Kim, and
  Chung}]{alleman-etal-2021-syntactic}
Matteo Alleman, Jonathan Mamou, Miguel A~Del~Rio, Hanlin Tang, Yoon Kim, and
  SueYeon Chung. 2021.
\newblock \href {https://doi.org/10.18653/v1/2021.repl4nlp-1.27} {Syntactic
  perturbations reveal representational correlates of hierarchical phrase
  structure in pretrained language models}.
\newblock In \emph{Proceedings of the 6th Workshop on Representation Learning
  for NLP (RepL4NLP-2021)}, pages 263--276, Online. Association for
  Computational Linguistics.

\bibitem[{Bai et~al.(2021)Bai, Wang, Chen, Yang, Bai, Yu, and
  Tong}]{bai-etal-2021-syntax}
Jiangang Bai, Yujing Wang, Yiren Chen, Yaming Yang, Jing Bai, Jing Yu, and
  Yunhai Tong. 2021.
\newblock \href {https://aclanthology.org/2021.eacl-main.262} {Syntax-{BERT}:
  Improving pre-trained transformers with syntax trees}.
\newblock In \emph{Proceedings of the 16th Conference of the European Chapter
  of the Association for Computational Linguistics: Main Volume}, pages
  3011--3020, Online. Association for Computational Linguistics.

\bibitem[{Bailyn(2012)}]{bailyn2012syntax}
John~F Bailyn. 2012.
\newblock \emph{{The Syntax of Russian}}.
\newblock Cambridge University Press.

\bibitem[{Belinkov(2021)}]{belinkov2021probing}
Yonatan Belinkov. 2021.
\newblock Probing classifiers: Promises, shortcomings, and alternatives.
\newblock \emph{arXiv preprint arXiv:2102.12452}.

\bibitem[{Belinkov and Glass(2019)}]{belinkov-glass-2019-analysis}
Yonatan Belinkov and James Glass. 2019.
\newblock \href {https://doi.org/10.1162/tacl_a_00254} {Analysis methods in
  neural language processing: A survey}.
\newblock \emph{Transactions of the Association for Computational Linguistics},
  7:49--72.

\bibitem[{B{\"o}rjars et~al.(2003)B{\"o}rjars, Engdahl, Andr{\'e}asson, Butt,
  and King}]{borjars2003subject}
Kersti B{\"o}rjars, Elisabet Engdahl, Maia Andr{\'e}asson, Miriam Butt, and
  Tracy~Holloway King. 2003.
\newblock {Subject and Object Positions in Swedish}.
\newblock In \emph{Proceedings of the LFG03 Conference}, pages 43--58.
  Citeseer.

\bibitem[{Chu(1965)}]{chu1965shortest}
Yoeng-Jin Chu. 1965.
\newblock {On the Shortest Arborescence of a Directed Graph}.
\newblock \emph{Scientia Sinica}, 14:1396--1400.

\bibitem[{Clark et~al.(2019)Clark, Khandelwal, Levy, and
  Manning}]{clark-etal-2019-bert}
Kevin Clark, Urvashi Khandelwal, Omer Levy, and Christopher~D. Manning. 2019.
\newblock \href {https://doi.org/10.18653/v1/W19-4828} {What does {BERT} look
  at? an analysis of {BERT}{'}s attention}.
\newblock In \emph{Proceedings of the 2019 ACL Workshop BlackboxNLP: Analyzing
  and Interpreting Neural Networks for NLP}, pages 276--286, Florence, Italy.
  Association for Computational Linguistics.

\bibitem[{Clouatre et~al.(2021)Clouatre, Parthasarathi, Zouaq, and
  Chandar}]{clouatre2021demystifying}
Louis Clouatre, Prasanna Parthasarathi, Amal Zouaq, and Sarath Chandar. 2021.
\newblock {Demystifying Neural Language Models' Insensitivity to Word-Order}.
\newblock \emph{arXiv preprint arXiv:2107.13955}.

\bibitem[{Conneau et~al.(2018)Conneau, Kruszewski, Lample, Barrault, and
  Baroni}]{conneau-etal-2018-cram}
Alexis Conneau, German Kruszewski, Guillaume Lample, Lo{\"\i}c Barrault, and
  Marco Baroni. 2018.
\newblock \href {https://doi.org/10.18653/v1/P18-1198} {What you can cram into
  a single {\$}{\&}!{\#}* vector: Probing sentence embeddings for linguistic
  properties}.
\newblock In \emph{Proceedings of the 56th Annual Meeting of the Association
  for Computational Linguistics (Volume 1: Long Papers)}, pages 2126--2136,
  Melbourne, Australia. Association for Computational Linguistics.

\bibitem[{Devlin et~al.(2019)Devlin, Chang, Lee, and
  Toutanova}]{devlin-etal-2019-bert}
Jacob Devlin, Ming-Wei Chang, Kenton Lee, and Kristina Toutanova. 2019.
\newblock \href {https://doi.org/10.18653/v1/N19-1423} {{BERT}: Pre-training of
  deep bidirectional transformers for language understanding}.
\newblock In \emph{Proceedings of the 2019 Conference of the North {A}merican
  Chapter of the Association for Computational Linguistics: Human Language
  Technologies, Volume 1 (Long and Short Papers)}, pages 4171--4186,
  Minneapolis, Minnesota. Association for Computational Linguistics.

\bibitem[{Dufter et~al.(2021)Dufter, Schmitt, and
  Sch{\"u}tze}]{dufter2021position}
Philipp Dufter, Martin Schmitt, and Hinrich Sch{\"u}tze. 2021.
\newblock {Position Information in Transformers: An Overview}.
\newblock \emph{arXiv preprint arXiv:2102.11090}.

\bibitem[{Edmonds(1968)}]{edmonds1968optimum}
Jack Edmonds. 1968.
\newblock {Optimum branchings}.
\newblock \emph{Mathematics and the Decision Sciences, Part}, 1(335-345):25.

\bibitem[{Eger et~al.(2020)Eger, Daxenberger, and
  Gurevych}]{eger-etal-2020-probe}
Steffen Eger, Johannes Daxenberger, and Iryna Gurevych. 2020.
\newblock \href {https://doi.org/10.18653/v1/2020.conll-1.8} {How to probe
  sentence embeddings in low-resource languages: On structural design choices
  for probing task evaluation}.
\newblock In \emph{Proceedings of the 24th Conference on Computational Natural
  Language Learning}, pages 108--118, Online. Association for Computational
  Linguistics.

\bibitem[{Ettinger(2020)}]{ettinger-2020-bert}
Allyson Ettinger. 2020.
\newblock \href {https://doi.org/10.1162/tacl_a_00298} {What {BERT} is not:
  Lessons from a new suite of psycholinguistic diagnostics for language
  models}.
\newblock \emph{Transactions of the Association for Computational Linguistics},
  8:34--48.

\bibitem[{Futrell et~al.(2018)Futrell, Wilcox, Morita, and
  Levy}]{futrell2018rnns}
Richard Futrell, Ethan Wilcox, Takashi Morita, and Roger Levy. 2018.
\newblock {RNNs as Psycholinguistic Subjects: Syntactic State and Grammatical
  Dependency}.
\newblock \emph{arXiv preprint arXiv:1809.01329}.

\bibitem[{Futrell et~al.(2019)Futrell, Wilcox, Morita, Qian, Ballesteros, and
  Levy}]{futrell-etal-2019-neural}
Richard Futrell, Ethan Wilcox, Takashi Morita, Peng Qian, Miguel Ballesteros,
  and Roger Levy. 2019.
\newblock \href {https://doi.org/10.18653/v1/N19-1004} {Neural language models
  as psycholinguistic subjects: Representations of syntactic state}.
\newblock In \emph{Proceedings of the 2019 Conference of the North {A}merican
  Chapter of the Association for Computational Linguistics: Human Language
  Technologies, Volume 1 (Long and Short Papers)}, pages 32--42, Minneapolis,
  Minnesota. Association for Computational Linguistics.

\bibitem[{Gauthier et~al.(2020)Gauthier, Hu, Wilcox, Qian, and
  Levy}]{gauthier2020syntaxgym}
Jon Gauthier, Jennifer Hu, Ethan Wilcox, Peng Qian, and Roger Levy. 2020.
\newblock {SyntaxGym: An Online Platform for Targeted Evaluation of Language
  Models}.
\newblock In \emph{Proceedings of the 58th Annual Meeting of the Association
  for Computational Linguistics: System Demonstrations}, pages 70--76.

\bibitem[{Ginter et~al.(2017)Ginter, Haji{\v c}, Luotolahti, Straka, and
  Zeman}]{ginter@conll}
Filip Ginter, Jan Haji{\v c}, Juhani Luotolahti, Milan Straka, and Daniel
  Zeman. 2017.
\newblock \href {http://hdl.handle.net/11234/1-1989} {{CoNLL} 2017 shared task
  - automatically annotated raw texts and word embeddings}.
\newblock {LINDAT}/{CLARIAH}-{CZ} digital library at the Institute of Formal
  and Applied Linguistics ({{\'U}FAL}), Faculty of Mathematics and Physics,
  Charles University.

\bibitem[{Goldberg(2019)}]{goldberg2019assessing}
Yoav Goldberg. 2019.
\newblock {Assessing BERT's syntactic abilities}.
\newblock \emph{arXiv preprint arXiv:1901.05287}.

\bibitem[{Gupta et~al.(2021)Gupta, Kvernadze, and Srikumar}]{gupta2021bert}
Ashim Gupta, Giorgi Kvernadze, and Vivek Srikumar. 2021.
\newblock {BERT \& Family Eat Word Salad: Experiments with Text Understanding}.
\newblock In \emph{Proceedings of the AAAI Conference on Artificial
  Intelligence}, volume~35, pages 12946--12954.

\bibitem[{Hessel and Schofield(2021)}]{hessel-schofield-2021-effective}
Jack Hessel and Alexandra Schofield. 2021.
\newblock \href {https://doi.org/10.18653/v1/2021.acl-short.27} {How effective
  is {BERT} without word ordering? implications for language understanding and
  data privacy}.
\newblock In \emph{Proceedings of the 59th Annual Meeting of the Association
  for Computational Linguistics and the 11th International Joint Conference on
  Natural Language Processing (Volume 2: Short Papers)}, pages 204--211,
  Online. Association for Computational Linguistics.

\bibitem[{Hewitt and Liang(2019)}]{hewitt-liang-2019-designing}
John Hewitt and Percy Liang. 2019.
\newblock \href {https://doi.org/10.18653/v1/D19-1275} {Designing and
  interpreting probes with control tasks}.
\newblock In \emph{Proceedings of the 2019 Conference on Empirical Methods in
  Natural Language Processing and the 9th International Joint Conference on
  Natural Language Processing (EMNLP-IJCNLP)}, pages 2733--2743, Hong Kong,
  China. Association for Computational Linguistics.

\bibitem[{Hewitt and Manning(2019)}]{hewitt-manning-2019-structural}
John Hewitt and Christopher~D. Manning. 2019.
\newblock \href {https://doi.org/10.18653/v1/N19-1419} {{A} structural probe
  for finding syntax in word representations}.
\newblock In \emph{Proceedings of the 2019 Conference of the North {A}merican
  Chapter of the Association for Computational Linguistics: Human Language
  Technologies, Volume 1 (Long and Short Papers)}, pages 4129--4138,
  Minneapolis, Minnesota. Association for Computational Linguistics.

\bibitem[{Hill et~al.(2016)Hill, Cho, and Korhonen}]{hill-etal-2016-learning}
Felix Hill, Kyunghyun Cho, and Anna Korhonen. 2016.
\newblock \href {https://doi.org/10.18653/v1/N16-1162} {Learning distributed
  representations of sentences from unlabelled data}.
\newblock In \emph{Proceedings of the 2016 Conference of the North {A}merican
  Chapter of the Association for Computational Linguistics: Human Language
  Technologies}, pages 1367--1377, San Diego, California. Association for
  Computational Linguistics.

\bibitem[{Htut et~al.(2019)Htut, Phang, Bordia, and Bowman}]{htut2019attention}
Phu~Mon Htut, Jason Phang, Shikha Bordia, and Samuel~R Bowman. 2019.
\newblock {Do Attention heads in BERT track syntactic dependencies?}
\newblock \emph{arXiv preprint arXiv:1911.12246}.

\bibitem[{Jawahar et~al.(2019)Jawahar, Sagot, and Seddah}]{jawahar2019does}
Ganesh Jawahar, Beno{\^\i}t Sagot, and Djam{\'e} Seddah. 2019.
\newblock {What Does BERT Learn about the Structure of Language?}
\newblock In \emph{Proceedings of the 57th Annual Meeting of the Association
  for Computational Linguistics}, pages 3651--3657.

\bibitem[{Jo and Myaeng(2020)}]{jo2020roles}
Jae-young Jo and Sung-Hyon Myaeng. 2020.
\newblock {Roles and Utilization of Attention Heads in Transformer-based Neural
  Language Models}.
\newblock In \emph{Proceedings of the 58th Annual Meeting of the Association
  for Computational Linguistics}, pages 3404--3417.

\bibitem[{Khandelwal et~al.(2018)Khandelwal, He, Qi, and
  Jurafsky}]{khandelwal-etal-2018-sharp}
Urvashi Khandelwal, He~He, Peng Qi, and Dan Jurafsky. 2018.
\newblock \href {https://doi.org/10.18653/v1/P18-1027} {Sharp nearby, fuzzy far
  away: How neural language models use context}.
\newblock In \emph{Proceedings of the 56th Annual Meeting of the Association
  for Computational Linguistics (Volume 1: Long Papers)}, pages 284--294,
  Melbourne, Australia. Association for Computational Linguistics.

\bibitem[{Kim et~al.(2020)Kim, Choi, Edmiston, and Lee}]{kim2020pre}
Taeuk Kim, Jihun Choi, Daniel Edmiston, and Sang-goo Lee. 2020.
\newblock Are pre-trained language models aware of phrases? simple but strong
  baselines for grammar induction.
\newblock \emph{arXiv preprint arXiv:2002.00737}.

\bibitem[{Klein and Manning(2004)}]{klein-manning-2004-corpus}
Dan Klein and Christopher Manning. 2004.
\newblock \href {https://doi.org/10.3115/1218955.1219016} {Corpus-based
  induction of syntactic structure: Models of dependency and constituency}.
\newblock In \emph{Proceedings of the 42nd Annual Meeting of the Association
  for Computational Linguistics ({ACL}-04)}, pages 478--485, Barcelona, Spain.

\bibitem[{Lau et~al.(2020)Lau, Armendariz, Lappin, Purver, and
  Shu}]{lau-etal-2020-furiously}
Jey~Han Lau, Carlos Armendariz, Shalom Lappin, Matthew Purver, and Chang Shu.
  2020.
\newblock \href {https://doi.org/10.1162/tacl_a_00315} {How furiously can
  colorless green ideas sleep? sentence acceptability in context}.
\newblock \emph{Transactions of the Association for Computational Linguistics},
  8:296--310.

\bibitem[{Lau et~al.(2017)Lau, Clark, and Lappin}]{lau2017grammaticality}
Jey~Han Lau, Alexander Clark, and Shalom Lappin. 2017.
\newblock {Grammaticality, Acceptability, and Probability: A Probabilistic View
  of Linguistic Lnowledge}.
\newblock \emph{Cognitive science}, 41(5):1202--1241.

\bibitem[{Lewis et~al.(2020)Lewis, Liu, Goyal, Ghazvininejad, Mohamed, Levy,
  Stoyanov, and Zettlemoyer}]{lewis-etal-2020-bart}
Mike Lewis, Yinhan Liu, Naman Goyal, Marjan Ghazvininejad, Abdelrahman Mohamed,
  Omer Levy, Veselin Stoyanov, and Luke Zettlemoyer. 2020.
\newblock \href {https://doi.org/10.18653/v1/2020.acl-main.703} {{BART}:
  Denoising sequence-to-sequence pre-training for natural language generation,
  translation, and comprehension}.
\newblock In \emph{Proceedings of the 58th Annual Meeting of the Association
  for Computational Linguistics}, pages 7871--7880, Online. Association for
  Computational Linguistics.

\bibitem[{Lin et~al.(2019)Lin, Tan, and Frank}]{lin-etal-2019-open}
Yongjie Lin, Yi~Chern Tan, and RoBERT Frank. 2019.
\newblock \href {https://doi.org/10.18653/v1/W19-4825} {Open sesame: Getting
  inside {BERT}{'}s linguistic knowledge}.
\newblock In \emph{Proceedings of the 2019 ACL Workshop BlackboxNLP: Analyzing
  and Interpreting Neural Networks for NLP}, pages 241--253, Florence, Italy.
  Association for Computational Linguistics.

\bibitem[{Liu et~al.(2019{\natexlab{a}})Liu, Gardner, Belinkov, Peters, and
  Smith}]{liu-etal-2019-linguistic}
Nelson~F. Liu, Matt Gardner, Yonatan Belinkov, Matthew~E. Peters, and Noah~A.
  Smith. 2019{\natexlab{a}}.
\newblock \href {https://doi.org/10.18653/v1/N19-1112} {Linguistic knowledge
  and transferability of contextual representations}.
\newblock In \emph{Proceedings of the 2019 Conference of the North {A}merican
  Chapter of the Association for Computational Linguistics: Human Language
  Technologies, Volume 1 (Long and Short Papers)}, pages 1073--1094,
  Minneapolis, Minnesota. Association for Computational Linguistics.

\bibitem[{Liu et~al.(2020{\natexlab{a}})Liu, Kusner, and
  Blunsom}]{liu2020survey}
Qi~Liu, Matt~J Kusner, and Phil Blunsom. 2020{\natexlab{a}}.
\newblock A survey on contextual embeddings.
\newblock \emph{arXiv preprint arXiv:2003.07278}.

\bibitem[{Liu et~al.(2020{\natexlab{b}})Liu, Gu, Goyal, Li, Edunov,
  Ghazvininejad, Lewis, and Zettlemoyer}]{liu-etal-2020-multilingual-denoising}
Yinhan Liu, Jiatao Gu, Naman Goyal, Xian Li, Sergey Edunov, Marjan
  Ghazvininejad, Mike Lewis, and Luke Zettlemoyer. 2020{\natexlab{b}}.
\newblock \href {https://doi.org/10.1162/tacl_a_00343} {Multilingual denoising
  pre-training for neural machine translation}.
\newblock \emph{Transactions of the Association for Computational Linguistics},
  8:726--742.

\bibitem[{Liu et~al.(2019{\natexlab{b}})Liu, Ott, Goyal, Du, Joshi, Chen, Levy,
  Lewis, Zettlemoyer, and Stoyanov}]{liu2019roberta}
Yinhan Liu, Myle Ott, Naman Goyal, Jingfei Du, Mandar Joshi, Danqi Chen, Omer
  Levy, Mike Lewis, Luke Zettlemoyer, and Veselin Stoyanov. 2019{\natexlab{b}}.
\newblock {RoBERTa: a Robustly Optimized BERT Pre-training Approach}.
\newblock \emph{arXiv preprint arXiv:1907.11692}.

\bibitem[{Liu et~al.(2021)Liu, Winata, Cahyawijaya, Madotto, Lin, and
  Fung}]{liu2021importance}
Zihan Liu, Genta~I Winata, Samuel Cahyawijaya, Andrea Madotto, Zhaojiang Lin,
  and Pascale Fung. 2021.
\newblock {On the Importance of Word Order Information in Cross-lingual
  Sequence Labeling}.
\newblock In \emph{Proceedings of the AAAI Conference on Artificial
  Intelligence}, volume~35, pages 13461--13469.

\bibitem[{Luo et~al.(2021)Luo, Kulmizev, and Mao}]{luo2021positional}
Ziyang Luo, Artur Kulmizev, and Xiaoxi Mao. 2021.
\newblock Positional artefacts propagate through masked language model
  embeddings.

\bibitem[{Maudslay and Cotterell(2021)}]{maudslay2021syntactic}
Rowan~Hall Maudslay and Ryan Cotterell. 2021.
\newblock {Do Syntactic Probes Probe Syntax? Experiments with Jabberwocky
  Probing}.
\newblock In \emph{Proceedings of the 2021 Conference of the North American
  Chapter of the Association for Computational Linguistics: Human Language
  Technologies}, pages 124--131.

\bibitem[{Miaschi et~al.(2020)Miaschi, Brunato, Dell{'}Orletta, and
  Venturi}]{miaschi-etal-2020-linguistic}
Alessio Miaschi, Dominique Brunato, Felice Dell{'}Orletta, and Giulia Venturi.
  2020.
\newblock \href {https://doi.org/10.18653/v1/2020.coling-main.65} {Linguistic
  profiling of a neural language model}.
\newblock In \emph{Proceedings of the 28th International Conference on
  Computational Linguistics}, pages 745--756, Barcelona, Spain (Online).
  International Committee on Computational Linguistics.

\bibitem[{Nie et~al.(2019)Nie, Wang, and Bansal}]{nie2019analyzing}
Yixin Nie, Yicheng Wang, and Mohit Bansal. 2019.
\newblock {Analyzing Compositionality-Sensitivity of NLI Models}.
\newblock In \emph{Proceedings of the AAAI Conference on Artificial
  Intelligence}, volume~33, pages 6867--6874.

\bibitem[{O'Connor and Andreas(2021)}]{o2021context}
Joe O'Connor and Jacob Andreas. 2021.
\newblock {What Context Features Can Transformer Language Models Use?}
\newblock \emph{arXiv preprint arXiv:2106.08367}.

\bibitem[{Panda et~al.(2021)Panda, Agrawal, Ha, and
  Bloch}]{panda-etal-2021-shuffled}
Subhadarshi Panda, Anjali Agrawal, Jeewon Ha, and Benjamin Bloch. 2021.
\newblock \href {https://doi.org/10.18653/v1/2021.naacl-srw.12} {Shuffled-token
  detection for refining pre-trained {R}o{BERT}a}.
\newblock In \emph{Proceedings of the 2021 Conference of the North American
  Chapter of the Association for Computational Linguistics: Student Research
  Workshop}, pages 88--93, Online. Association for Computational Linguistics.

\bibitem[{Pedregosa et~al.(2011)Pedregosa, Varoquaux, Gramfort, Michel,
  Thirion, Grisel, Blondel, Prettenhofer, Weiss, Dubourg
  et~al.}]{pedregosa2011scikit}
Fabian Pedregosa, Ga{\"e}l Varoquaux, Alexandre Gramfort, Vincent Michel,
  BERTrand Thirion, Olivier Grisel, Mathieu Blondel, Peter Prettenhofer, Ron
  Weiss, Vincent Dubourg, et~al. 2011.
\newblock Scikit-learn: Machine learning in python.
\newblock \emph{the Journal of machine Learning research}, 12:2825--2830.

\bibitem[{Pham et~al.(2020)Pham, Bui, Mai, and Nguyen}]{pham2020out}
Thang~M Pham, Trung Bui, Long Mai, and Anh Nguyen. 2020.
\newblock {Out of Order: How Important is the Sequential Order of Words in a
  Sentence in Natural Language Understanding Tasks?}
\newblock \emph{arXiv preprint arXiv:2012.15180}.

\bibitem[{Prince(1988)}]{prince1988pragmatic}
Ellen~F Prince. 1988.
\newblock {On Pragmatic Change: the Borrowing of Discourse Functions}.
\newblock \emph{Journal of pragmatics}, 12(5-6):505--518.

\bibitem[{Radford et~al.(2019)Radford, Wu, Child, Luan, Amodei, and
  Sutskever}]{radford2019language}
Alec Radford, Jeff Wu, Rewon Child, David Luan, Dario Amodei, and Ilya
  Sutskever. 2019.
\newblock {Language Models are Unsupervised Multitask Learners}.

\bibitem[{Raganato and Tiedemann(2018)}]{raganato-tiedemann-2018-analysis}
Alessandro Raganato and J{\"o}rg Tiedemann. 2018.
\newblock \href {https://doi.org/10.18653/v1/W18-5431} {An analysis of encoder
  representations in transformer-based machine translation}.
\newblock In \emph{Proceedings of the 2018 {EMNLP} Workshop {B}lackbox{NLP}:
  Analyzing and Interpreting Neural Networks for {NLP}}, pages 287--297,
  Brussels, Belgium. Association for Computational Linguistics.

\bibitem[{Ravishankar et~al.(2019)Ravishankar, {\O}vrelid, and
  Velldal}]{ravishankar-etal-2019-probing}
Vinit Ravishankar, Lilja {\O}vrelid, and Erik Velldal. 2019.
\newblock \href {https://doi.org/10.18653/v1/W19-4318} {Probing multilingual
  sentence representations with {X}-probe}.
\newblock In \emph{Proceedings of the 4th Workshop on Representation Learning
  for NLP (RepL4NLP-2019)}, pages 156--168, Florence, Italy. Association for
  Computational Linguistics.

\bibitem[{Rogers et~al.(2020)Rogers, Kovaleva, and
  Rumshisky}]{rogers-etal-2020-primer}
Anna Rogers, Olga Kovaleva, and Anna Rumshisky. 2020.
\newblock \href {https://doi.org/10.1162/tacl_a_00349} {A primer in
  {BERT}ology: What we know about how {BERT} works}.
\newblock \emph{Transactions of the Association for Computational Linguistics},
  8:842--866.

\bibitem[{Rosa and Mareček(2019)}]{rosa2019inducing}
Rudolf Rosa and David Mareček. 2019.
\newblock \href {http://arxiv.org/abs/1906.11511} {{Inducing Syntactic Trees
  from BERT Representations}}.

\bibitem[{Sachan et~al.(2021)Sachan, Zhang, Qi, and
  Hamilton}]{sachan-etal-2021-syntax}
Devendra Sachan, Yuhao Zhang, Peng Qi, and William~L. Hamilton. 2021.
\newblock \href {https://aclanthology.org/2021.eacl-main.228} {Do syntax trees
  help pre-trained transformers extract information?}
\newblock In \emph{Proceedings of the 16th Conference of the European Chapter
  of the Association for Computational Linguistics: Main Volume}, pages
  2647--2661, Online. Association for Computational Linguistics.

\bibitem[{{\c{S}}ahin and Steedman(2018)}]{sahin-steedman-2018-data}
G{\"o}zde~G{\"u}l {\c{S}}ahin and Mark Steedman. 2018.
\newblock \href {https://doi.org/10.18653/v1/D18-1545} {Data augmentation via
  dependency tree morphing for low-resource languages}.
\newblock In \emph{Proceedings of the 2018 Conference on Empirical Methods in
  Natural Language Processing}, pages 5004--5009, Brussels, Belgium.
  Association for Computational Linguistics.

\bibitem[{{\c{S}}ahin and Steedman(2019)}]{csahin2019data}
G{\"o}zde~G{\"u}l {\c{S}}ahin and Mark Steedman. 2019.
\newblock Data augmentation via dependency tree morphing for low-resource
  languages.
\newblock \emph{arXiv preprint arXiv:1903.09460}.

\bibitem[{{\c{S}}ahin et~al.(2020){\c{S}}ahin, Vania, Kuznetsov, and
  Gurevych}]{sahin-etal-2020-linspector}
G{\"o}zde~G{\"u}l {\c{S}}ahin, Clara Vania, Ilia Kuznetsov, and Iryna Gurevych.
  2020.
\newblock \href {https://doi.org/10.1162/coli_a_00376} {{LINSPECTOR}:
  Multilingual probing tasks for word representations}.
\newblock \emph{Computational Linguistics}, 46(2):335--385.

\bibitem[{Salazar et~al.(2020)Salazar, Liang, Nguyen, and
  Kirchhoff}]{salazar-etal-2020-masked}
Julian Salazar, Davis Liang, Toan~Q. Nguyen, and Katrin Kirchhoff. 2020.
\newblock \href {https://doi.org/10.18653/v1/2020.acl-main.240} {Masked
  language model scoring}.
\newblock In \emph{Proceedings of the 58th Annual Meeting of the Association
  for Computational Linguistics}, pages 2699--2712, Online. Association for
  Computational Linguistics.

\bibitem[{Sankar et~al.(2019)Sankar, Subramanian, Pal, Chandar, and
  Bengio}]{sankar2019neural}
Chinnadhurai Sankar, Sandeep Subramanian, Christopher Pal, Sarath Chandar, and
  Yoshua Bengio. 2019.
\newblock Do neural dialog systems use the conversation history effectively? an
  empirical study.
\newblock In \emph{Proceedings of the 57th Annual Meeting of the Association
  for Computational Linguistics}, pages 32--37.

\bibitem[{Si et~al.(2019)Si, Wang, Kan, and Jiang}]{si2019does}
Chenglei Si, Shuohang Wang, Min-Yen Kan, and Jing Jiang. 2019.
\newblock What does bert learn from multiple-choice reading comprehension
  datasets?
\newblock \emph{arXiv preprint arXiv:1910.12391}.

\bibitem[{Sinha et~al.(2021)Sinha, Jia, Hupkes, Pineau, Williams, and
  Kiela}]{DBLP:journals/corr/abs-2104-06644}
Koustuv Sinha, Robin Jia, Dieuwke Hupkes, Joelle Pineau, Adina Williams, and
  Douwe Kiela. 2021.
\newblock \href {http://arxiv.org/abs/2104.06644} {Masked language modeling and
  the distributional hypothesis: Order word matters pre-training for little}.
\newblock \emph{CoRR}, abs/2104.06644.

\bibitem[{Sinha et~al.(2020)Sinha, Parthasarathi, Pineau, and
  Williams}]{sinha2020unnatural}
Koustuv Sinha, Prasanna Parthasarathi, Joelle Pineau, and Adina Williams. 2020.
\newblock Unnatural language inference.
\newblock \emph{arXiv preprint arXiv:2101.00010}.

\bibitem[{Sugawara et~al.(2020)Sugawara, Stenetorp, Inui, and
  Aizawa}]{sugawara2020assessing}
Saku Sugawara, Pontus Stenetorp, Kentaro Inui, and Akiko Aizawa. 2020.
\newblock {Assessing the Benchmarking Capacity of Machine Reading Comprehension
  Datasets}.
\newblock In \emph{Proceedings of the AAAI Conference on Artificial
  Intelligence}, volume~34, pages 8918--8927.

\bibitem[{Tao et~al.(2021)Tao, Gao, Li, Feng, Zhao, and
  Yan}]{tao-etal-2021-learning}
Chongyang Tao, Shen Gao, Juntao Li, Yansong Feng, Dongyan Zhao, and Rui Yan.
  2021.
\newblock \href {https://doi.org/10.18653/v1/2021.naacl-main.134} {Learning to
  organize a bag of words into sentences with neural networks: An empirical
  study}.
\newblock In \emph{Proceedings of the 2021 Conference of the North American
  Chapter of the Association for Computational Linguistics: Human Language
  Technologies}, pages 1682--1691, Online. Association for Computational
  Linguistics.

\bibitem[{Tenney et~al.(2018)Tenney, Xia, Chen, Wang, Poliak, McCoy, Kim,
  Van~Durme, Bowman, Das et~al.}]{tenney2018you}
Ian Tenney, Patrick Xia, Berlin Chen, Alex Wang, Adam Poliak, R~Thomas McCoy,
  Najoung Kim, Benjamin Van~Durme, Samuel~R Bowman, Dipanjan Das, et~al. 2018.
\newblock What do you learn from context? probing for sentence structure in
  contextualized word representations.
\newblock In \emph{International Conference on Learning Representations}.

\bibitem[{Vaswani et~al.(2017)Vaswani, Shazeer, Parmar, Uszkoreit, Jones,
  Gomez, Kaiser, and Polosukhin}]{vaswani2017attention}
Ashish Vaswani, Noam Shazeer, Niki Parmar, Jakob Uszkoreit, Llion Jones,
  Aidan~N Gomez, {\L}ukasz Kaiser, and Illia Polosukhin. 2017.
\newblock Attention is all you need.
\newblock In \emph{Advances in neural information processing systems}, pages
  5998--6008.

\bibitem[{Vilares et~al.(2020)Vilares, Strzyz, Søgaard, and
  Gómez-Rodríguez}]{vilares2020parsing}
David Vilares, Michalina Strzyz, Anders Søgaard, and Carlos Gómez-Rodríguez.
  2020.
\newblock \href {https://doi.org/10.1609/aaai.v34i05.6446} {{Parsing as
  Pretraining}}.
\newblock 34:9114--9121.

\bibitem[{Volodina et~al.(2021)Volodina, Mohammed, and
  Klezl}]{volodina2021dalaj}
Elena Volodina, Yousuf~Ali Mohammed, and Julia Klezl. 2021.
\newblock {DaLAJ-a Dataset for Linguistic Acceptability Judgments for Swedish:
  Format, Baseline, Sharing}.
\newblock \emph{arXiv preprint arXiv:2105.06681}.

\bibitem[{Wang et~al.(2018)Wang, Singh, Michael, Hill, Levy, and
  Bowman}]{wang-etal-2018-glue}
Alex Wang, Amanpreet Singh, Julian Michael, Felix Hill, Omer Levy, and Samuel
  Bowman. 2018.
\newblock \href {https://doi.org/10.18653/v1/W18-5446} {{GLUE}: A multi-task
  benchmark and analysis platform for natural language understanding}.
\newblock In \emph{Proceedings of the 2018 {EMNLP} Workshop {B}lackbox{NLP}:
  Analyzing and Interpreting Neural Networks for {NLP}}, pages 353--355,
  Brussels, Belgium. Association for Computational Linguistics.

\bibitem[{Wang et~al.(2020)Wang, Shang, Lioma, Jiang, Yang, Liu, and
  Simonsen}]{wang2020position}
Benyou Wang, Lifeng Shang, Christina Lioma, Xin Jiang, Hao Yang, Qun Liu, and
  Jakob~Grue Simonsen. 2020.
\newblock {On Position Embeddings in BERT}.
\newblock In \emph{International Conference on Learning Representations}.

\bibitem[{Wang et~al.(2019)Wang, Bi, Yan, Wu, Bao, Xia, Peng, and
  Si}]{wang2019structbert}
Wei Wang, Bin Bi, Ming Yan, Chen Wu, Zuyi Bao, Jiangnan Xia, Liwei Peng, and
  Luo Si. 2019.
\newblock Structbert: incorporating language structures into pre-training for
  deep language understanding.
\newblock \emph{arXiv preprint arXiv:1908.04577}.

\bibitem[{Wang and Chen(2020)}]{wang-chen-2020-position}
Yu-An Wang and Yun-Nung Chen. 2020.
\newblock \href {https://doi.org/10.18653/v1/2020.emnlp-main.555} {What do
  position embeddings learn? an empirical study of pre-trained language model
  positional encoding}.
\newblock In \emph{Proceedings of the 2020 Conference on Empirical Methods in
  Natural Language Processing (EMNLP)}, pages 6840--6849, Online. Association
  for Computational Linguistics.

\bibitem[{Warstadt and Bowman(2019)}]{DBLP:journals/corr/abs-1901-03438}
Alex Warstadt and Samuel~R. Bowman. 2019.
\newblock \href {http://arxiv.org/abs/1901.03438} {Grammatical analysis of
  pretrained sentence encoders with acceptability judgments}.
\newblock \emph{CoRR}, abs/1901.03438.

\bibitem[{Warstadt et~al.(2020)Warstadt, Parrish, Liu, Mohananey, Peng, Wang,
  and Bowman}]{warstadt-etal-2020-blimp-benchmark}
Alex Warstadt, Alicia Parrish, Haokun Liu, Anhad Mohananey, Wei Peng, Sheng-Fu
  Wang, and Samuel~R. Bowman. 2020.
\newblock \href {https://doi.org/10.1162/tacl_a_00321} {{BL}i{MP}: The
  benchmark of linguistic minimal pairs for {E}nglish}.
\newblock \emph{Transactions of the Association for Computational Linguistics},
  8:377--392.

\bibitem[{Warstadt et~al.(2019)Warstadt, Singh, and
  Bowman}]{warstadt-etal-2019-neural}
Alex Warstadt, Amanpreet Singh, and Samuel~R. Bowman. 2019.
\newblock \href {https://doi.org/10.1162/tacl_a_00290} {Neural network
  acceptability judgments}.
\newblock \emph{Transactions of the Association for Computational Linguistics},
  7:625--641.

\bibitem[{Wolf et~al.(2020)Wolf, Debut, Sanh, Chaumond, Delangue, Moi, Cistac,
  Rault, Louf, Funtowicz, Davison, Shleifer, von Platen, Ma, Jernite, Plu, Xu,
  Le~Scao, Gugger, Drame, Lhoest, and Rush}]{wolf-etal-2020-transformers}
Thomas Wolf, Lysandre Debut, Victor Sanh, Julien Chaumond, Clement Delangue,
  Anthony Moi, Pierric Cistac, Tim Rault, Remi Louf, Morgan Funtowicz, Joe
  Davison, Sam Shleifer, Patrick von Platen, Clara Ma, Yacine Jernite, Julien
  Plu, Canwen Xu, Teven Le~Scao, Sylvain Gugger, Mariama Drame, Quentin Lhoest,
  and Alexander Rush. 2020.
\newblock \href {https://doi.org/10.18653/v1/2020.emnlp-demos.6} {Transformers:
  State-of-the-art natural language processing}.
\newblock In \emph{Proceedings of the 2020 Conference on Empirical Methods in
  Natural Language Processing: System Demonstrations}, pages 38--45, Online.
  Association for Computational Linguistics.

\bibitem[{Wu et~al.(2020)Wu, Chen, Kao, and Liu}]{wu-etal-2020-perturbed}
Zhiyong Wu, Yun Chen, Ben Kao, and Qun Liu. 2020.
\newblock \href {https://doi.org/10.18653/v1/2020.acl-main.383} {Perturbed
  masking: Parameter-free probing for analyzing and interpreting {BERT}}.
\newblock In \emph{Proceedings of the 58th Annual Meeting of the Association
  for Computational Linguistics}, pages 4166--4176, Online. Association for
  Computational Linguistics.

\bibitem[{Xiang et~al.(2021)Xiang, Yang, Li, Warstadt, and
  Kann}]{xiang-etal-2021-climp}
Beilei Xiang, Changbing Yang, Yu~Li, Alex Warstadt, and Katharina Kann. 2021.
\newblock \href {https://aclanthology.org/2021.eacl-main.242} {{CL}i{MP}: A
  benchmark for {C}hinese language model evaluation}.
\newblock In \emph{Proceedings of the 16th Conference of the European Chapter
  of the Association for Computational Linguistics: Main Volume}, pages
  2784--2790, Online. Association for Computational Linguistics.

\end{thebibliography}
\bibliographystyle{acl_natbib}

\onecolumn
\setcounter{section}{0}
\setcounter{figure}{0}

\section*{\centering {Appendix}}

\section{Dataset Statistics}
\label{app:stata} 

\begin{table*}[h!]
\centering
\begin{tabular}{c|c|c|c|c}
\toprule
\textbf{} & \textbf{Language} & \textbf{NgramShift} & \textbf{ClauseShift} & \textbf{RandomShift} \\
\midrule


\textbf{num. tokens} & 
    \begin{tabular}{@{}c@{}c@{}}\textbf{Ru} \\ \textbf{En} \\ \textbf{Sv}\end{tabular}
  & \begin{tabular}{@{}c@{}c@{}} 105.8k \\ 128.5k \\ 134.1k \end{tabular} & \begin{tabular}{@{}c@{}c@{}}  199.7k \\ 198.6k \\ 192.9k \end{tabular} & \begin{tabular}{@{}c@{}c@{}} 95.6k \\ 111.1k \\ 100.7k \end{tabular}
  
  \\ \midrule
  
\textbf{unique tokens} & 
    \begin{tabular}{@{}c@{}c@{}}\textbf{Ru} \\ \textbf{En} \\ \textbf{Sv}\end{tabular}
  & \begin{tabular}{@{}c@{}c@{}} 25.2k  \\ 19.2k \\ 23.2k \end{tabular} & \begin{tabular}{@{}c@{}c@{}}  46.1k \\ 25.1k \\ 25.7k \end{tabular} & \begin{tabular}{@{}c@{}c@{}} 27.8k \\ 22.8k \\ 17.8k \end{tabular}
  
  \\ \midrule
  
\textbf{tokens / sentence} & 
    \begin{tabular}{@{}c@{}c@{}}\textbf{Ru} \\ \textbf{En} \\ \textbf{Sv}\end{tabular}
  & \begin{tabular}{@{}c@{}c@{}} 10.9 \\ 12.9 \\ 13.4 \end{tabular} & \begin{tabular}{@{}c@{}c@{}}  19.9 \\ 19.9 \\ 19.3 \end{tabular} & \begin{tabular}{@{}c@{}c@{}} 10.5 \\ 11.1 \\ 10.1 \end{tabular}
  
\\ \bottomrule

\end{tabular}
\caption{A brief statistics of the controlled perturbation datasets. Languages: \textbf{Ru}=Russian, \textbf{En}=English, \textbf{Sv}=Swedish.}
\label{tab:tab_stata}
\end{table*}

\clearpage

\section{Parameter-free Probing}
\label{app:pfp}

\begin{figure*}[h!]
  \centering
  \includegraphics[width=.8\textwidth]{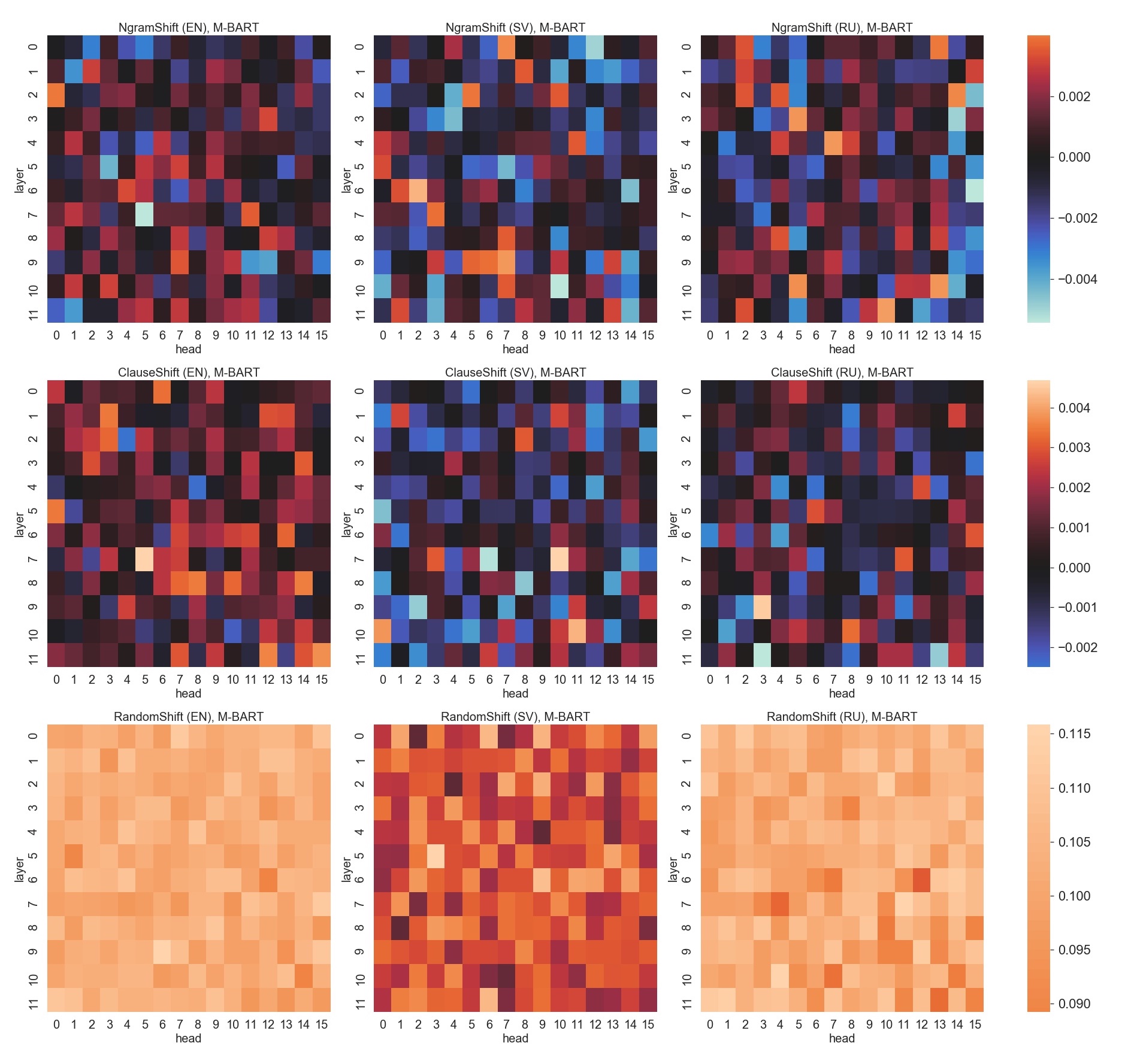}
  \caption{The task-wise heatmaps depicting the $\delta$ UUAS scores by M-BART for each language. Method=\textbf{Self-Attention Probing}. PE=\textbf{absolute}. X-axis=Attention head index. Y-axis=Layer index. Tasks: \textbf{NgramShift} (top); \textbf{ClauseShift} (middle); \textbf{RandomShift} (bottom). Languages: \textbf{En}=English (left); \textbf{Sv}=Swedish (middle); \textbf{Ru}=Russian (right).}
  \label{fig:mbart_complexity}
\end{figure*}

\begin{figure*}[h!]
  \centering
  \includegraphics[width=.8\textwidth]{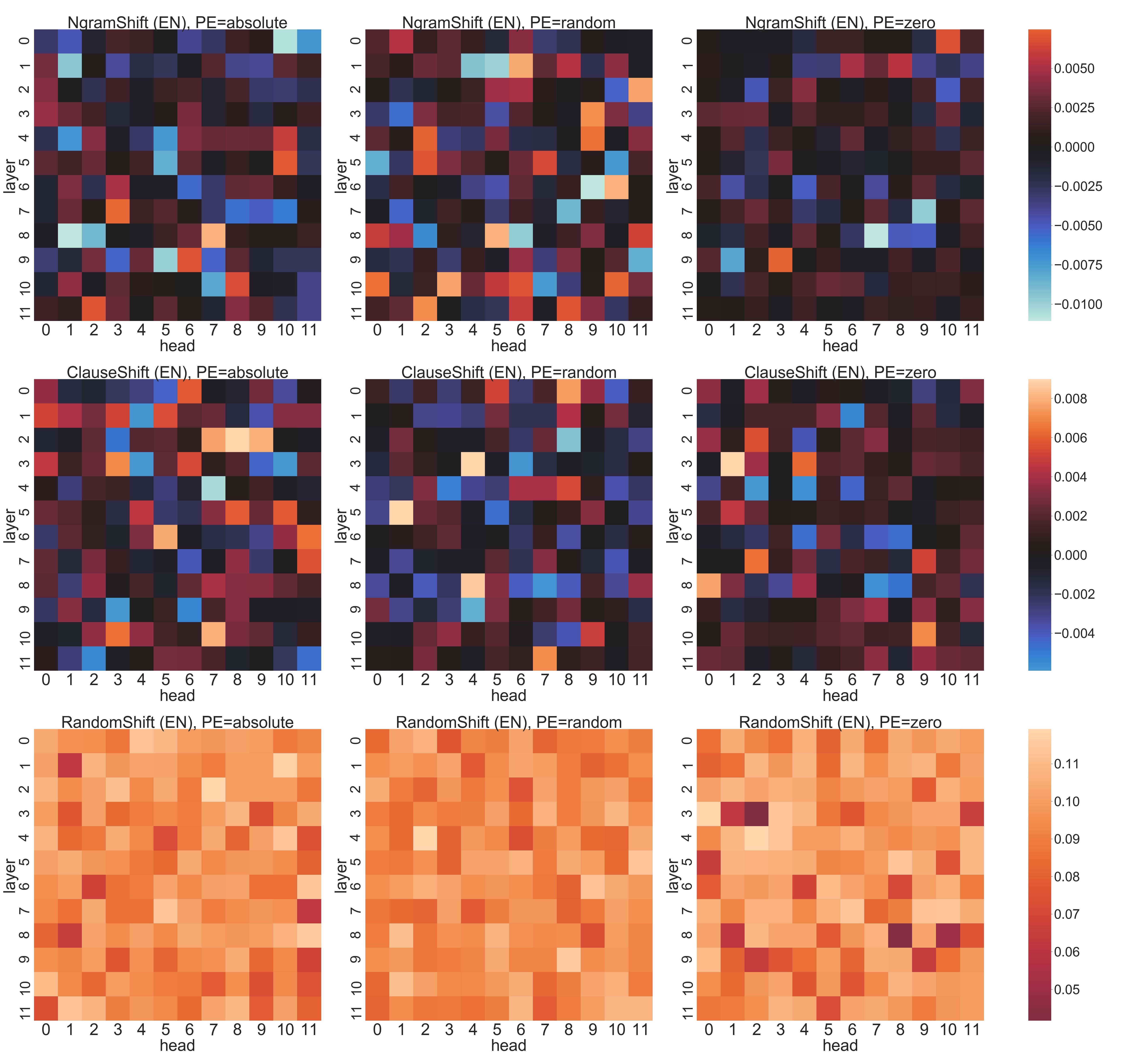}
  \caption{The task-wise heatmaps depicting the $\delta$ UUAS scores by M-BERT for each language. Method=\textbf{Self-Attention Probing}. PE: \textbf{absolute} (left); \textbf{random} (middle); \textbf{zero} (right). X-axis=Attention head index. Y-axis=Layer index. Tasks: \textbf{NgramShift} (top); \textbf{ClauseShift} (middle); \textbf{RandomShift} (bottom).}
  \label{fig:mbert_complexity_pe}
\end{figure*}

\begin{figure*}[t!]
  \centering
  \includegraphics[width=.85\textwidth]{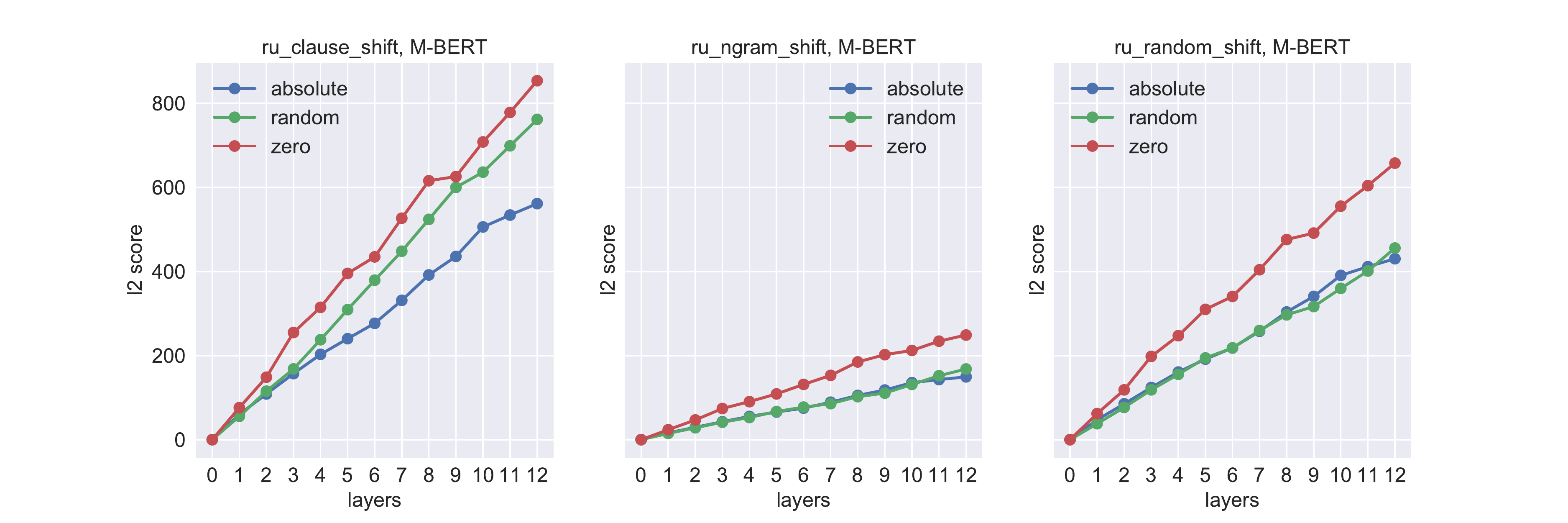}
  \caption{The Euclidean distance between the impact matrices computed by M-BERT with different PEs over each pair of sentences ($s$, $s'$) for Russian. The distances are averaged over attention heads at each layer. Method: \textbf{Token Perturbed Masking}. Tasks: \textbf{NgramShift} (left); \textbf{ClauseShift} (middle); \textbf{RandomShift} (right)}. 
  \label{fig:l2-perturbed-ru-mbert}
\end{figure*}

\begin{figure*}[t!]
  \centering
  \includegraphics[width=.85\textwidth]{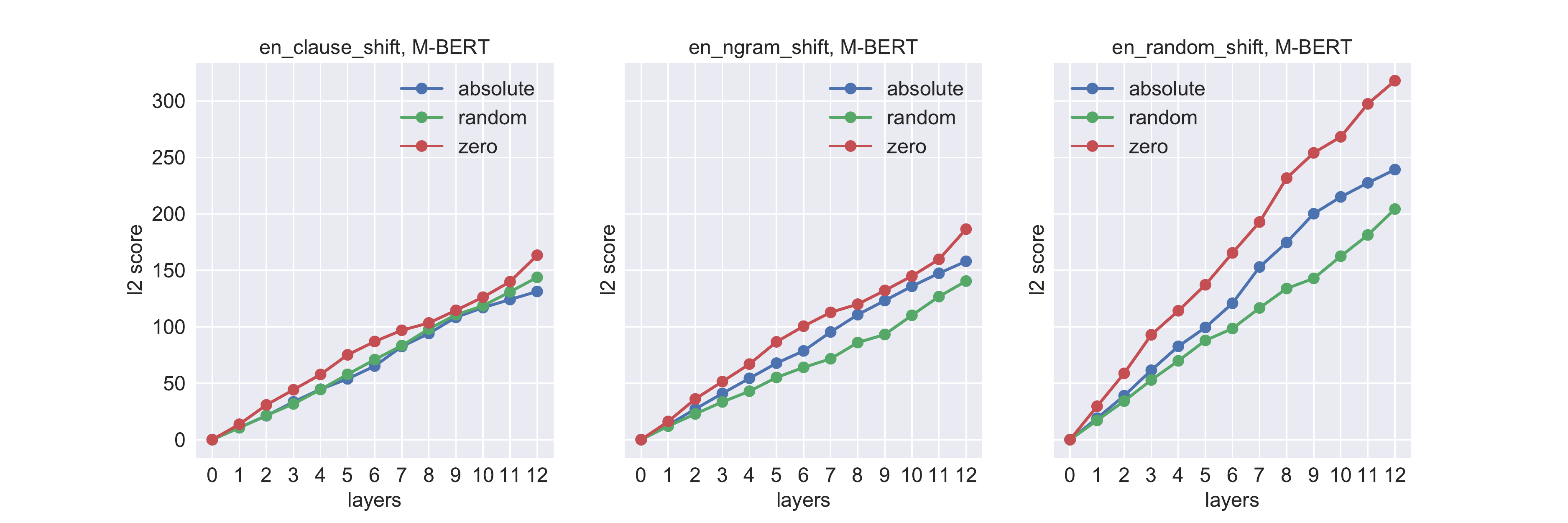}
  \caption{The Euclidean distance between the impact matrices computed by M-BERT with different PEs over each pair of sentences ($s$, $s'$) for English. The distances are averaged over attention heads at each layer. Method: \textbf{Token Perturbed Masking}. Tasks: \textbf{NgramShift} (left); \textbf{ClauseShift} (middle); \textbf{RandomShift} (right)}. 
  \label{fig:l2-perturbed-en-mbert}
\end{figure*}

\begin{figure*}[t!]
  \centering
  \includegraphics[width=.85\textwidth]{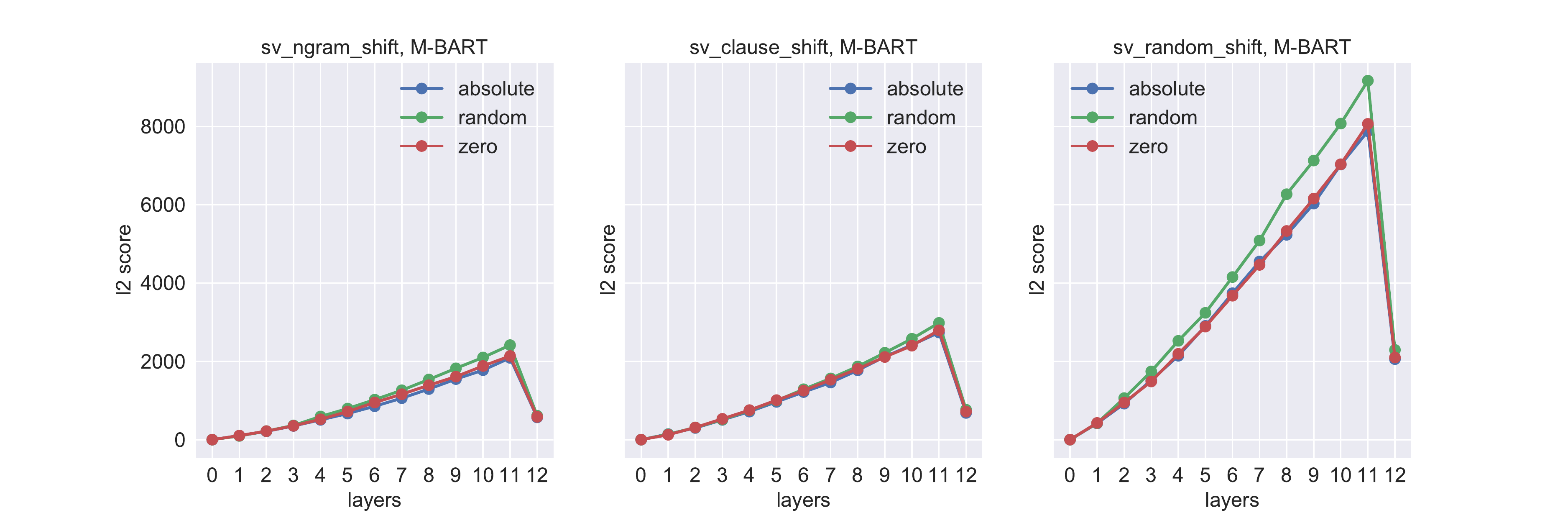}
  \caption{The Euclidean distance between the impact matrices computed by M-BART with different PEs over each pair of sentences ($s$, $s'$) for Swedish. The distances are averaged over attention heads at each layer. Method: \textbf{Token Perturbed Masking}. Tasks: \textbf{NgramShift} (left); \textbf{ClauseShift} (middle); \textbf{RandomShift} (right)}. 
  \label{fig:l2-perturbed-sv-mbart}
\end{figure*}

\begin{figure*}[h!]
  \centering
  \includegraphics[width=.85\textwidth]{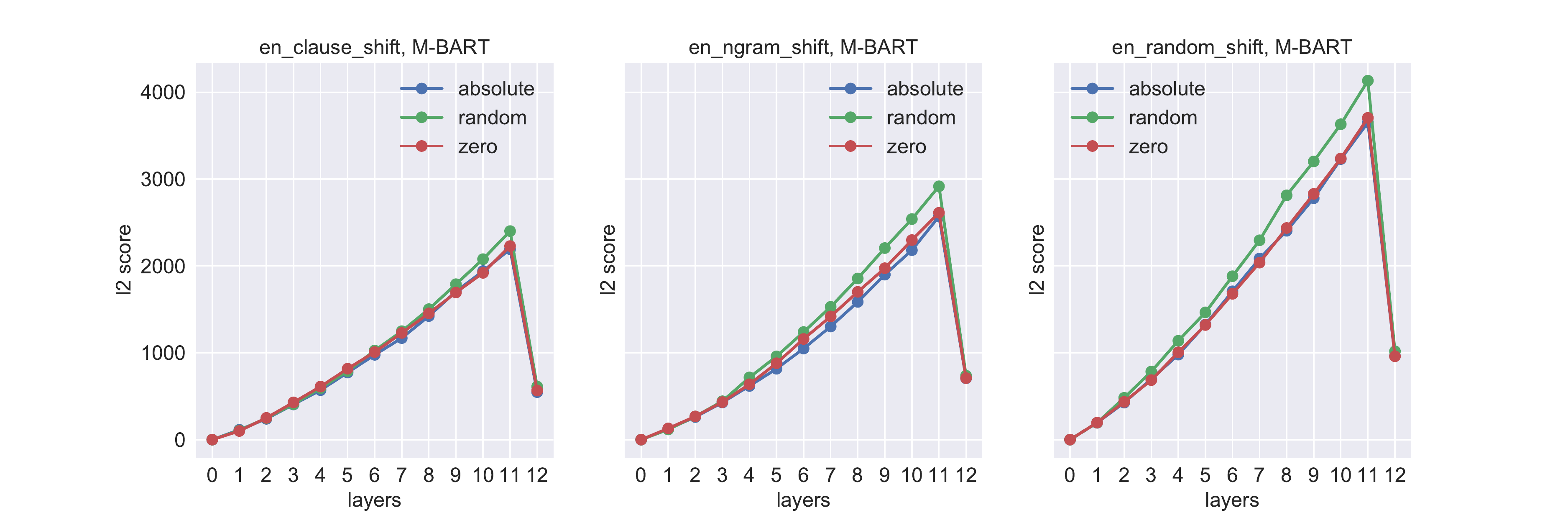}
  \caption{The Euclidean distance between the impact matrices computed by M-BART with different PEs over each pair of sentences ($s$, $s'$) for English. The distances are averaged over attention heads at each layer. Method: \textbf{Token Perturbed Masking}. Tasks: \textbf{NgramShift} (left); \textbf{ClauseShift} (middle); \textbf{RandomShift} (right)}. 
  \label{fig:l2-perturbed-en-mbart}
\end{figure*}

\begin{figure*}[h!]
  \centering
  \includegraphics[width=.85\textwidth]{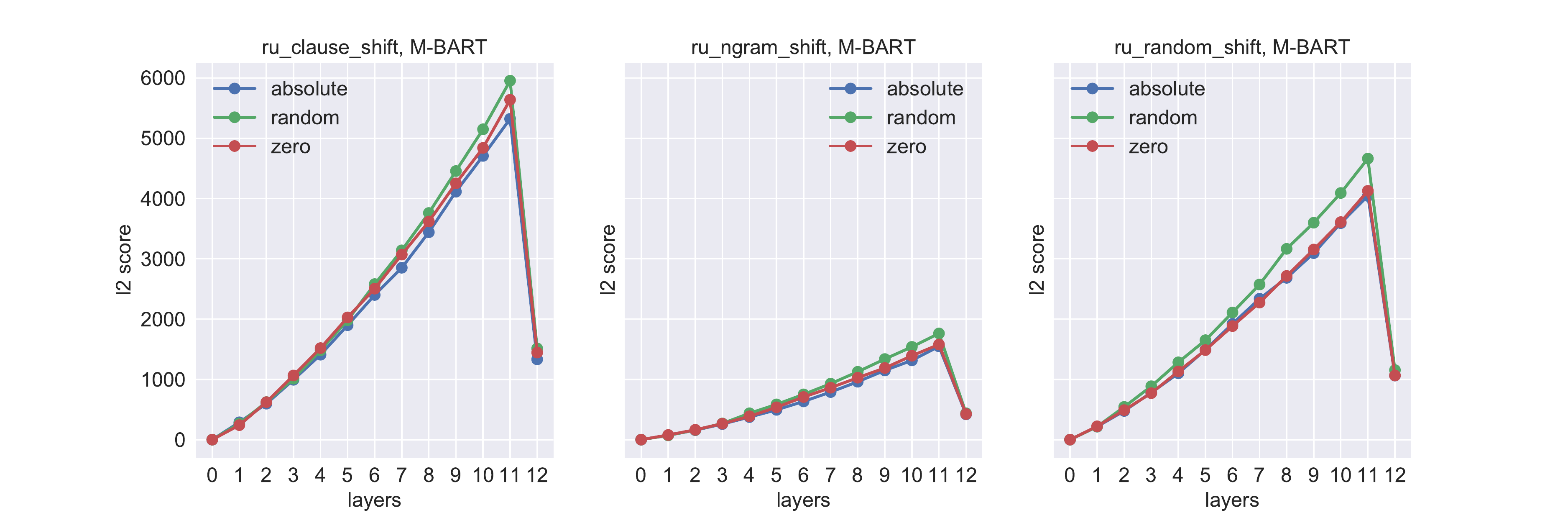}
  \caption{The Euclidean distance between the impact matrices computed by M-BART with different PEs over each pair of sentences ($s$, $s'$) for Russian. The distances are averaged over attention heads at each layer. Method: \textbf{Token Perturbed Masking}. Tasks: \textbf{NgramShift} (left); \textbf{ClauseShift} (middle); \textbf{RandomShift} (right)}. 
  \label{fig:l2-perturbed-ru-mbart}
\end{figure*}

\begin{table*}[ht!]
\centering
\begin{tabular}{c|c|c|c}
\toprule
\textbf{} & \textbf{Language} & \textbf{M-BERT} &\textbf{ M-BART} \\
\midrule

\textbf{NgramShift} & 
    \begin{tabular}{@{}c@{}c@{}}\textbf{En} \\ \textbf{Sv} \\ \textbf{Ru}\end{tabular}
  & \begin{tabular}{@{}c@{}c@{}} 0.32; 0.33 \\ 0.30; 0.31 \\0.36; 0.38 \end{tabular} & \begin{tabular}{@{}c@{}c@{}}  0.31; 0.32 \\ 0.30; 0.31\\  0.37; 0.38  \end{tabular}
  
  \\ \midrule
  
 \textbf{ClauseShift} & 
    \begin{tabular}{@{}c@{}c@{}}\textbf{En} \\ \textbf{Sv} \\ \textbf{Ru}\end{tabular}
  & \begin{tabular}{@{}c@{}c@{}} 0.20; 0.21 \\ 0.20; 0.21 \\ 0.20; 0.21 \end{tabular} & \begin{tabular}{@{}c@{}c@{}} 0.20; 0.21 \\ 0.20; 0.21\\ 0.20; 0.21 \end{tabular} 
  \\ \midrule
  
\textbf{RandomShift} & 
    \begin{tabular}{@{}c@{}c@{}}\textbf{En} \\ \textbf{Sv} \\ \textbf{Ru}\end{tabular}
  & \begin{tabular}{@{}c@{}c@{}} 0.37; 0.38  \\ 0.38; 0.41 \\ 0.39; 0.42 \end{tabular} & \begin{tabular}{@{}c@{}c@{}} 0.37; 0.37 \\0.37; 0.39 \\ 0.40; 0.41 \end{tabular} 
\\ \bottomrule

\end{tabular}
\caption{The UUAS scores by \textbf{Self-Attention Probing} method. The minimum and maximum values are given (min; max). Languages: \textbf{Ru}=Russian, \textbf{En}=English, \textbf{Sv}=Swedish.}
\label{tab:uuas-sap}
\end{table*}

\begin{table*}[t!]
\centering
\begin{tabular}{c|c|c|c}
\toprule
\textbf{} & \textbf{Language} & \textbf{M-BERT} & \textbf{M-BART} \\
\midrule

\textbf{NgramShift} & 
    \begin{tabular}{@{}c@{}c@{}}\textbf{En} \\ \textbf{Sv} \\ \textbf{Ru}\end{tabular}
  & \begin{tabular}{@{}c@{}c@{}} 0.36; 0.44 \\ 0.37; 0.46 \\0.4244; 0.52 \end{tabular} & \begin{tabular}{@{}c@{}c@{}}  0.31; 0.44 \\ 0.43; 0.53\\  0.46; 0.54  \end{tabular} 
  
  \\ \midrule
  
 \textbf{ClauseShift} & 
    \begin{tabular}{@{}c@{}c@{}}\textbf{En} \\ \textbf{Sv} \\ \textbf{Ru}\end{tabular}
  & \begin{tabular}{@{}c@{}c@{}} 0.23; 0.29 \\ 0.25; 0.33 \\ 0.23; 0.28 \end{tabular} & \begin{tabular}{@{}c@{}c@{}} 0.23; 0.29  \\ 0.24; 0.33\\ 0.22; 0.28 \end{tabular} 
  \\ \midrule
  
\textbf{RandomShift} & 
    \begin{tabular}{@{}c@{}c@{}}\textbf{En} \\ \textbf{Sv} \\ \textbf{Ru}\end{tabular}
  & \begin{tabular}{@{}c@{}c@{}} 0.39; 0.47  \\0.46; 0.53\\ 0.46; 0.54 \end{tabular} & \begin{tabular}{@{}c@{}c@{}} 0.39; 0.47 \\0.43; 0.53 \\ 0.43; 0.54 \end{tabular} 
\\ \bottomrule

\end{tabular}
\caption{The UUAS scores by \textbf{Token Perturbed Masking} probe. The minimum and maximum values are given (min; max). Languages: \textbf{Ru}=Russian, \textbf{En}=English, \textbf{Sv}=Swedish.}
\label{tab:uuas-tpm}
\end{table*}

\begin{figure*}[h!]
    \centering
    \begin{subfigure}[b]{0.49\linewidth}
    \centering
    \includegraphics[width=\linewidth]{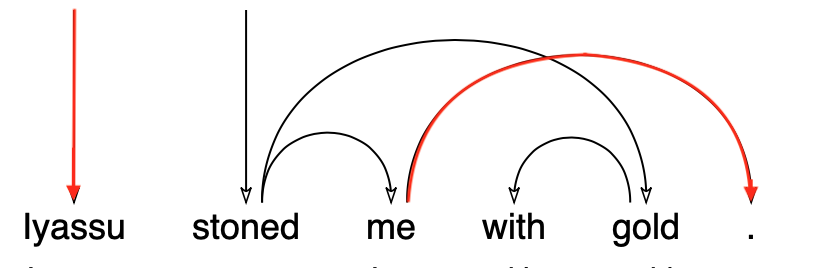} 
    \caption{\texttt{original}}
    \label{fig:tr-en-rshift-gr}
    \end{subfigure}
    \begin{subfigure}[b]{0.49\linewidth}
    \centering
    \includegraphics[width=\linewidth]{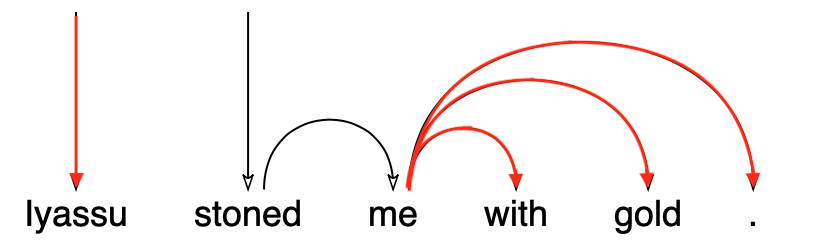}
    \caption{\texttt{perturbed}}
    \label{fig:tr-en-rshift-ungr}
    \end{subfigure}  
    \begin{subfigure}[b]{0.49\linewidth}
    \centering
    \includegraphics[width=\linewidth]{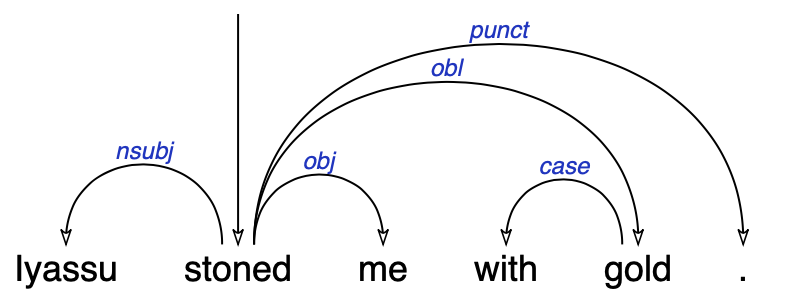}
    \caption{\texttt{gold}}
    \label{fig:tr-en-rshift-ud}
    \end{subfigure}
    \caption{Graphical representations of the syntactic trees inferred for the English sentence \textit{Iyassu stoned me with gold} and its perturbed version. \texttt{original}=the original sentence; \texttt{perturbed}=the perturbed version; \texttt{gold}=gold standard. Task=\textbf{RandomShift}. Model=\textbf{M-BERT} (Layer: 11; Head: 2). Method=\textbf{Self-Attention Probing}. The perturbation is underlined with red, and incorrectly assigned dependency heads are marked with red arrows.
    }
    \label{fig:tr-en}
\end{figure*}

\clearpage

\section{Representation Analysis}
\label{app:repr}

\begin{figure*}[h!]
    \centering
    \begin{subfigure}[b]{0.49\linewidth}
    \centering
    \includegraphics[width=\linewidth]{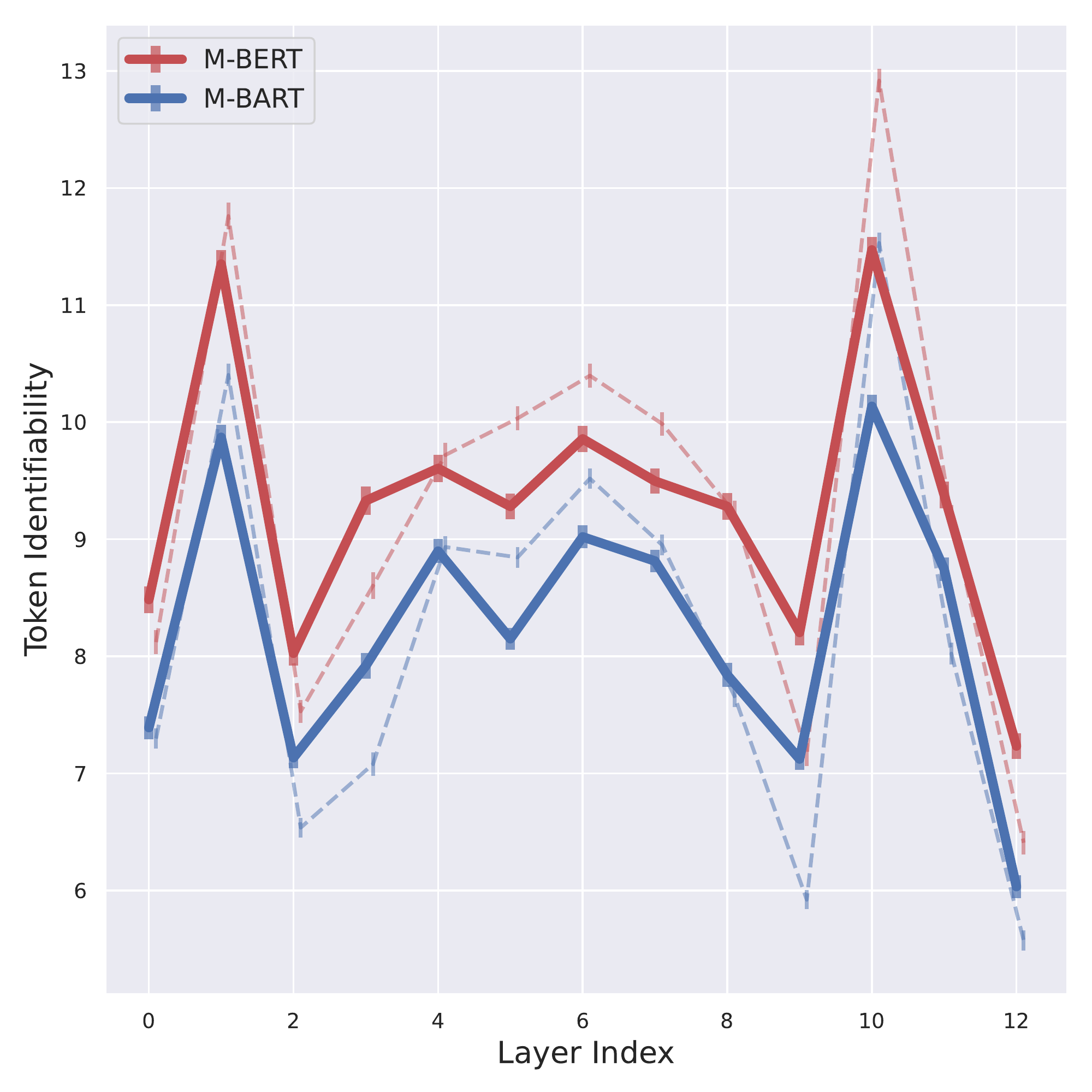}
    \caption{\textbf{NgramShift}}
    \label{fig:ti-ru-nshift}
    \end{subfigure}
    \begin{subfigure}[b]{0.49\linewidth}
    \centering
    \includegraphics[width=\linewidth]{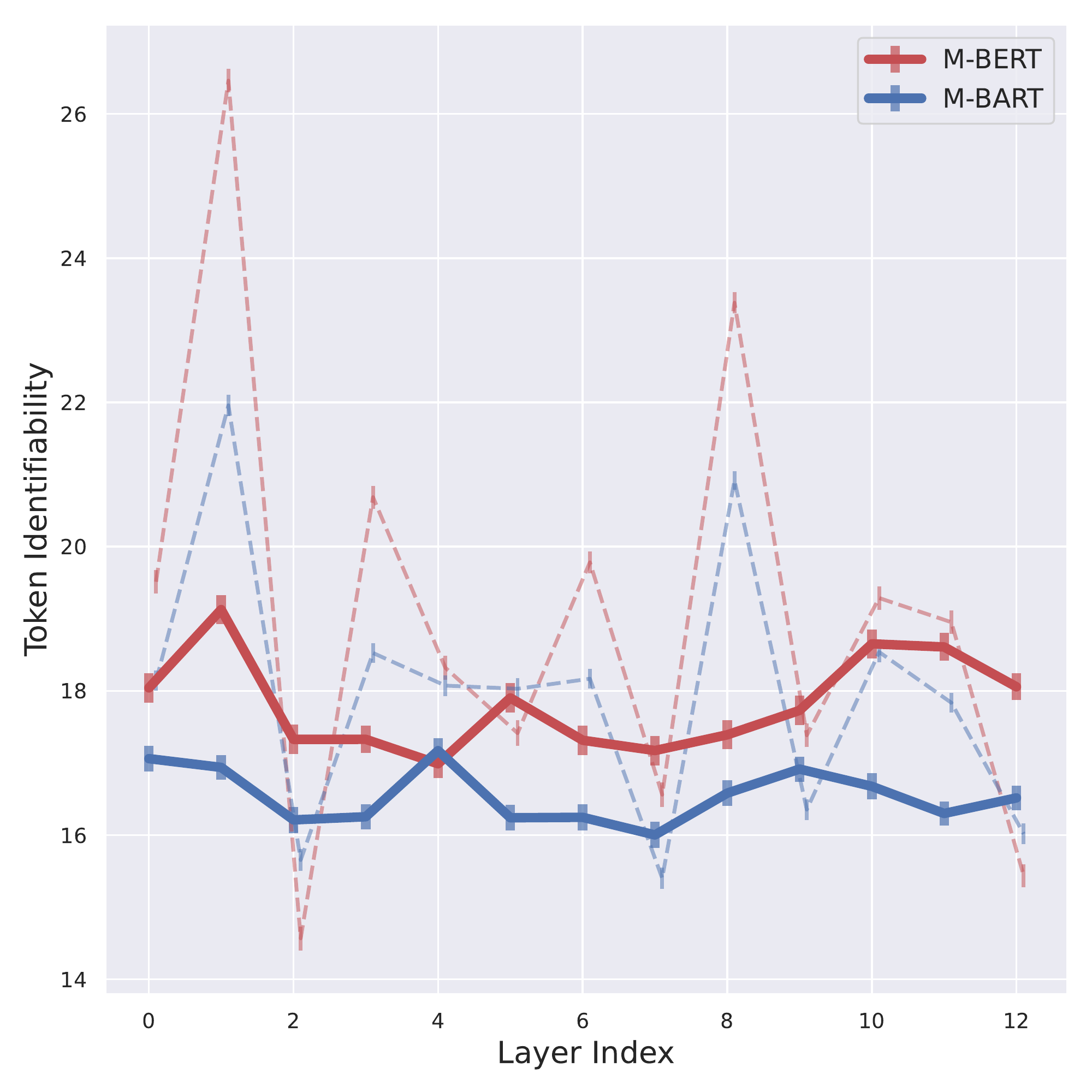}
    \caption{\textbf{ClauseShift}}
    \label{fig:ti-ru-cshift}
    \end{subfigure}    
    \caption{Token identifiability (TI) by layer for M-BERT and M-BART on the \textbf{NgramShift} (left) and \textbf{ClauseShift} (right) tasks for Russian. Dashed lines represent the scores computed over the intact sentences. X-axis=Layer index. Y-axis=TI.
    }
    \label{fig:ti-ru}

    \centering
    \includegraphics[width=\linewidth]{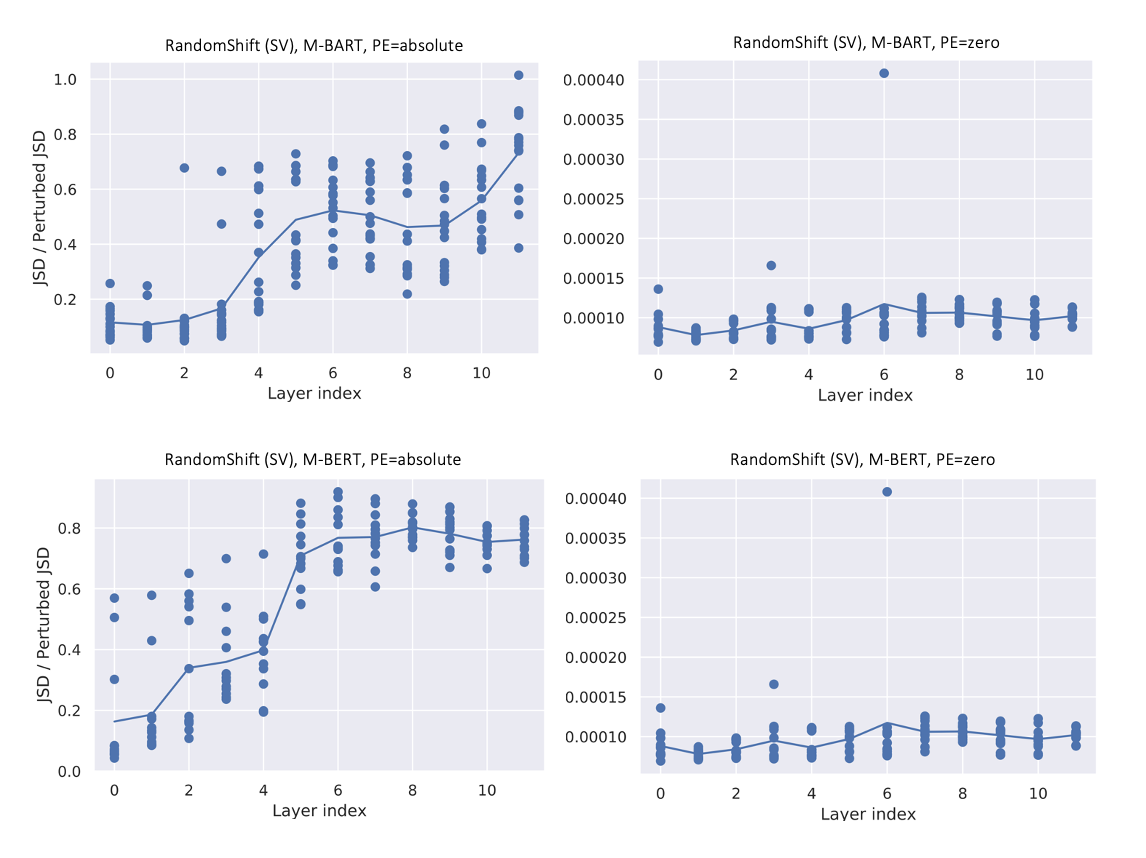}
    \caption{Self-Attention Distance (SAD) by layer for M-BART and M-BERT with absolute (left) and zeroed (right) positional embeddings on the \textbf{RandomShift} task for Swedish. X-axis=Layer index. Y-axis=SAD.
    }
    \label{fig:sad-sv}
\end{figure*}

\clearpage

\section{Acceptability Judgements}
\label{app:pppl}
\begin{figure*}[ht!]
  \centering
  \includegraphics[width=0.95\textwidth]{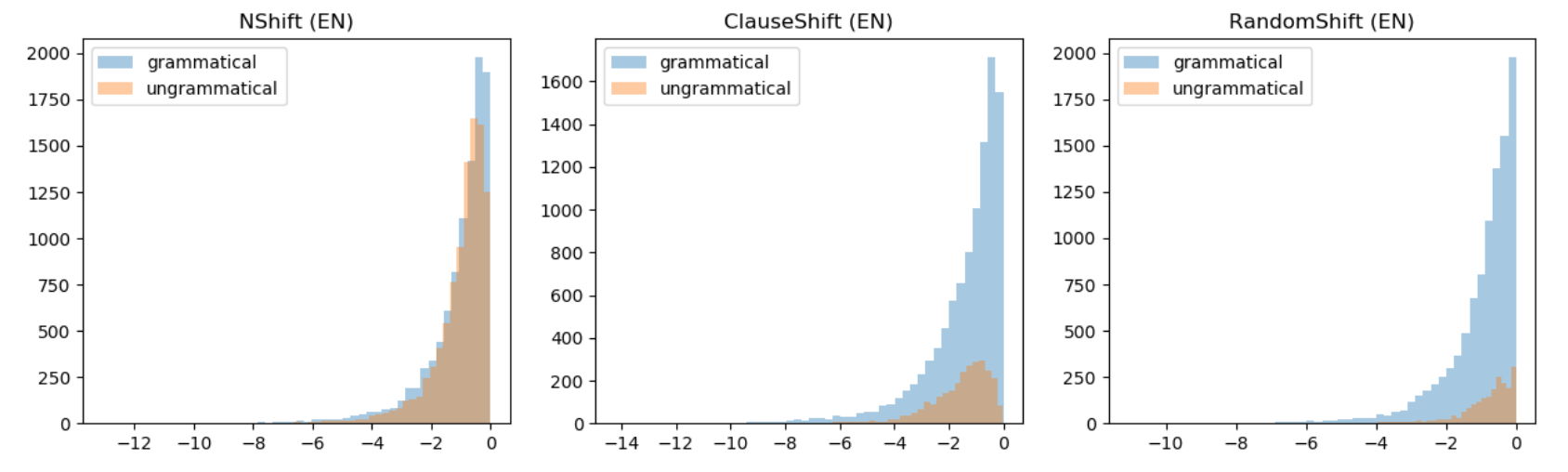}
  \caption{The \textit{MeanLP} distributions for the perturbed (ungrammatical) and intact (grammatical) sentences by M-BART. Tasks: \textbf{NgramShift} (left); \textbf{ClauseShift} (middle); \textbf{RandomShift} (right).}. 
  \label{fig:pppl-mbart}

  \centering
  \includegraphics[width=0.9\textwidth]{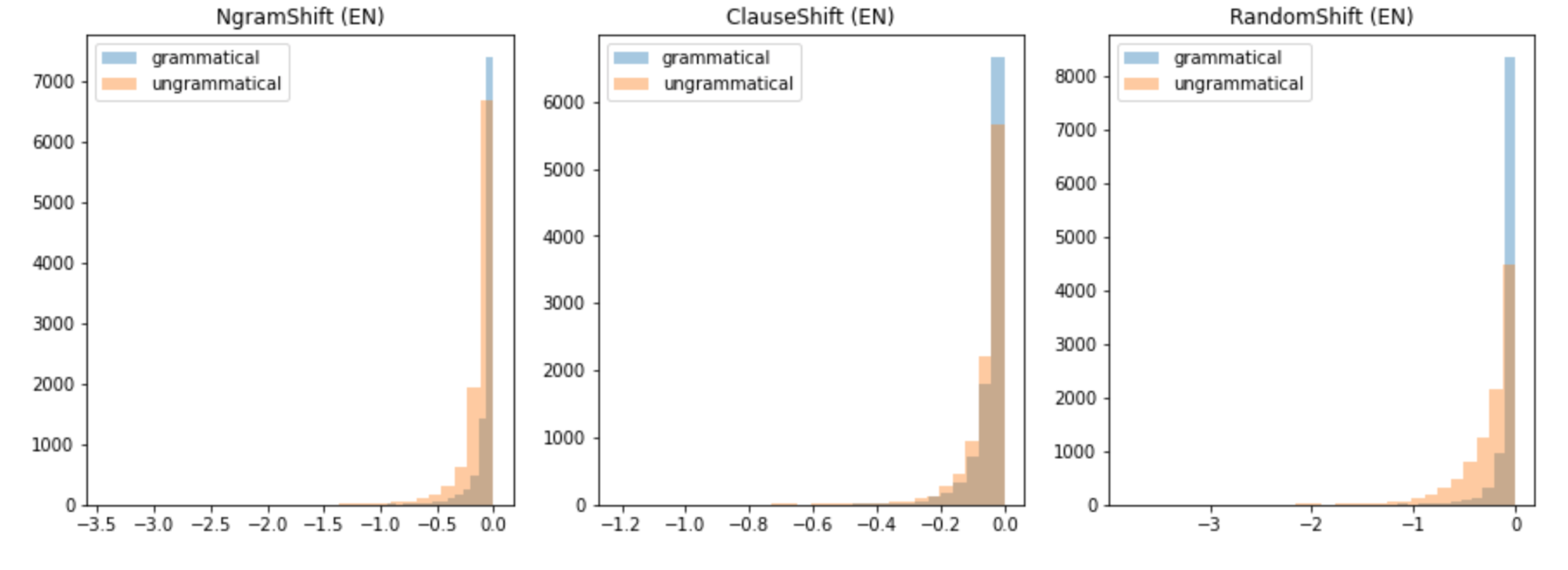}
  \caption{The \textit{MeanLP} distributions for the perturbed (ungrammatical) and intact (grammatical) sentences by M-BERT. Tasks: \textbf{NgramShift} (left); \textbf{ClauseShift} (middle); \textbf{RandomShift} (right).}. 
  \label{fig:pppl-mbert}
\end{figure*}

\end{document}